\documentclass[10pt]{wlscirep}
\usepackage[utf8]{inputenc}
\usepackage[T1]{fontenc}
\usepackage{multirow}
\usepackage{makecell}
\usepackage{booktabs}
\usepackage{bm}
\usepackage{subcaption}
\usepackage{caption}
\usepackage[table]{xcolor}

\definecolor{resnet-18}{HTML}{DCFCE7}
\definecolor{resnet-34}{HTML}{86EFAC}
\definecolor{resnet-50}{HTML}{4ADE80}
\definecolor{resnet-101}{HTML}{16A34A}

\title{Scaling Vision Models Does Not Consistently Improve Localisation-Based Explanation Quality}

\author[1*]{Mateusz Cedro}
\author[2]{Marcin Chlebus}
\affil[1]{University of Antwerp, Antwerp, 2000, Belgium}
\affil[2]{University of Warsaw, Warsaw, 00-241, Poland}

\affil[*]{mateusz.cedro@uantwerpen.be}


\begin{abstract}
Artificial intelligence models are increasingly scaled to improve predictive accuracy, yet it remains unclear whether scale improves the quality of post-hoc explanations. We investigate this relationship by evaluating 11 computer vision models representing increasing levels of depth and complexity within the ResNet, DenseNet, and Vision Transformer families, trained from scratch or pretrained, across three image datasets with ground-truth segmentation masks. For each model, we generate explanations using five post-hoc explainable AI methods and quantify mask alignment using two localisation metrics: Relevance Rank Accuracy~\cite{arras_ground_2022} and the proposed Dual-Polarity Precision, which measures positive attributions inside the class mask and negative attributions outside it. Across datasets and methods, increasing architectural depth and parameter count does not improve explanation quality in most statistical comparisons, and smaller models often match or exceed deeper variants. While pretraining typically improves predictive performance and increases the dependence of explanations on learned weights, it does not consistently increase localisation scores. We also observe scenarios in which models achieve strong predictive performance while localisation precision is near zero, suggesting that performance metrics alone may not indicate whether predictions are based on the annotated regions. These results indicate that larger models do not reliably provide higher-quality explanations, and that explainability should therefore be assessed explicitly during model selection for safety-sensitive deployments.

\end{abstract}

\begin{document}

\newcommand{\adjp}[2]{\ensuremath{#1^{#2}}}

\flushbottom
\maketitle
\thispagestyle{empty}

\section*{Introduction}
Deep neural networks show strong performance across many areas of artificial intelligence. However, their internal decision logic is often difficult to interpret, which raises concerns in settings such as healthcare and finance where transparency matters~\cite{morch_visualization_1995,baehrens_how_nodate,simonyan_deep_2014}. Explainable AI (XAI) addresses this issue by developing methods that provide insight into how models produce outputs across applications~\cite{lundberg_unified_2017,sundararajan_axiomatic_2017,arras_ground_2022,molnar2022,Dobrzycka31122025}.

In computer vision, Convolutional Neural Networks (CNNs) and Vision Transformers (ViT) reach high performance across different tasks, yet they are usually treated as black boxes because of their architectural complexity and large number of parameters~\cite{guidotti_survey_2019,ribeiro_why_2016,apley_visualizing_2020,dosovitskiy2020image}. Although images are directly interpretable to humans, it is often unclear which visual evidence drives a given prediction. Visual explanation methods, including saliency maps, aim to address this gap by identifying image regions that most influence the model output and providing a spatial account of model attention~\cite{simonyan_deep_2014}.

\begin{figure}[t!]
  \centering
  \includegraphics[width=0.8\textwidth]{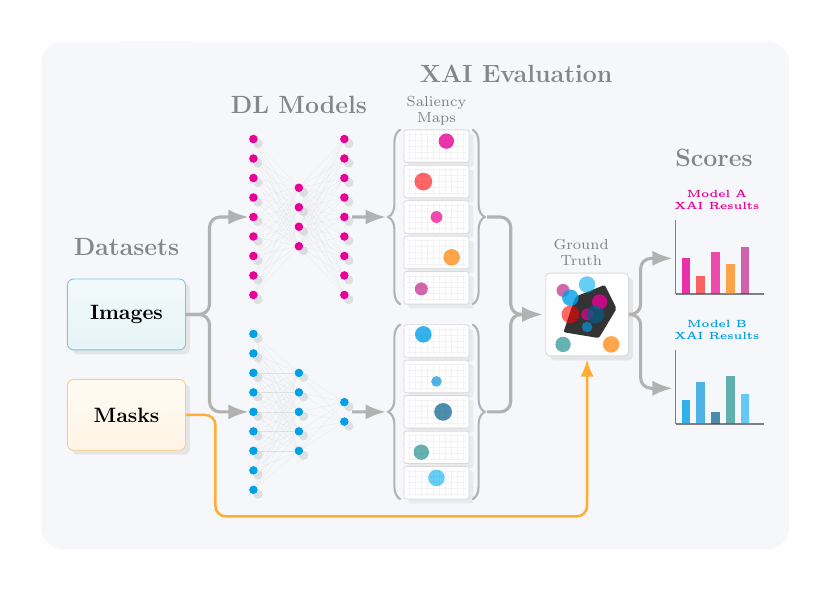}
  \caption{Schematic overview of the experimental methodology. Deep learning models are trained or fine-tuned on the image datasets and then used to generate saliency maps for test samples. The resulting explanations are compared with ground truth masks and assessed using Relevance Rank Accuracy~\cite{hedstrom2023quantus} and the proposed Dual-Polarity Precision to quantify the precision of the XAI outputs.}
  \label{fig:overview}
\end{figure}

\subsection*{Interpretable and Explainable AI}
Research on interpretable machine learning and explainable artificial intelligence (XAI) develops methods that make the behavior of deep learning models more understandable to human users~\cite{ribeiro_why_2016,samek_explainable_2019,molnar_interpretable_2020,cedro2024graphxain}. Interpretability describes the extent to which a system makes its reasoning understandable to humans~\cite{miller2019explanation}. Model assessment therefore often considers criteria beyond predictive accuracy, including whether explanations are coherent and reflect the model's behavior under controlled checks~\cite{adebayo_sanity_2020}. Interpretability is commonly described at two levels. Global interpretability characterizes general decision patterns across inputs, while local interpretability explains individual predictions~\cite{doshi-velez_towards_2017, molnar2022}. Together, these perspectives support bias analysis and decision justification in high-stakes settings.

A central topic in XAI is feature attribution, which assigns quantitative relevance to input features with respect to a model output~\cite{sundararajan_axiomatic_2017}. Methods such as DeepLIFT~\cite{shrikumar_learning_2019}, Layerwise Relevance Propagation~\cite{bach_pixel-wise_2015}, LIME~\cite{ribeiro_why_2016}, and Integrated Gradients~\cite{sundararajan_axiomatic_2017} estimate how inputs contribute to predictions. Game-theoretic approaches, including SHAP, derive feature contributions from cooperative game formulations~\cite{Shapley_1953,lundberg_unified_2017}. In computer vision, these attributions are typically presented as saliency or sensitivity maps, and approaches such as Grad-CAM and NoiseGrad aim to reduce noise and improve the spatial specificity of these visual explanations~\cite{smilkov_smoothgrad_2017,bykov_noisegrad_2022,selvaraju2017grad}. Nevertheless, establishing trust in these explanations requires careful evaluation of developed methods~\cite{hooker_benchmark_2019}. Validation of feature importance remains challenging because perturbation-based tests that remove or modify features can change the input distribution, which can confound the interpretation of the results~\cite{bach_pixel-wise_2015,sundararajan_axiomatic_2017}.

\subsection*{Explainable AI in Sensitive Imaging Applications}
AI is increasingly used in domains such as healthcare to support decisions including diagnosis and treatment planning, often through deep learning models trained on large labelled imaging datasets~\cite{lecun_deep_2015,holzinger_causability_2019}. As image-based systems move from rule-based approaches to data-driven methods, model complexity makes it harder for clinicians to understand the basis of individual predictions. This creates a need for methods that provide evidence for why a model produces a given output and that support informed review by domain experts~\cite{cabitza_unintended_2017,rajkomar_machine_2019}. XAI methods address this need by providing explanations that can be examined alongside model outputs, which helps users judge whether automated recommendations are reliable for the case at hand~\cite{katuwal_machine_nodate, che_interpretable_nodate, hinton_deep_2018}.

Using deep learning systems in practice also involves high computational and economic costs. Increasing model size can improve performance, but gains often decrease relative to the additional training and inference resources required~\cite{wu_wider_2016,zhaounderstanding}. These constraints motivate approaches that balance performance with compute and operational requirements in settings such as clinical workflows~\cite{brown_language_2020, menghani_efficient_2023,davenport_potential_2019,secinaro_role_2021}. During the COVID-19 pandemic, many studies develop models that predict disease status from chest X-rays~\cite{chowdhury_can_2020,degerli_covid-19_2021,rahman_exploring_2021,he_deep_2015}. In this context, saliency maps are widely used to provide spatial explanations, allowing clinicians to assess whether predictions rely on lung regions consistent with pathology rather than on spurious correlations or acquisition artifacts~\cite{showkat_efficacy_2022,saporta_benchmarking_2022}.

\subsection*{Influence of Model Scale on Performance and XAI Evaluations}\label{subsec3}
A common view in machine learning is that larger and more complex models achieve higher predictive accuracy~\cite{eigen_understanding_2014}. In practice, the relationship depends on the task, data regime, and training setup. Increasing depth can improve results, but larger architectures do not always outperform smaller ones. For example, in some settings, compact models such as ResNet-18 reach accuracy comparable to deeper variants~\cite{khan_evaluating_2018, sarwinda_deep_2021}. When datasets are limited, increasing capacity can also increase overfitting~\cite{brigato2021close}.

Empirical studies further report diminishing returns when complexity grows beyond a task-dependent range. Additional layers can only yield small gains relative to increased compute, and performance can stagnate or decrease under some conditions~\cite{eigen_understanding_2014, wu_wider_2016, guo_classification_2019}. These results motivate model selection procedures that consider both computational cost and performance requirements.

Scaling laws in deep learning indicate that increasing parameter count often improves predictive performance, but it is unclear whether the same trend holds for the fidelity of model explanations~\cite{lecun_deep_2015,hestness2017deep,bahri2024explaining}. A common assumption is that deeper models learn features with stronger semantic structure, which could yield explanations that align better with human interpretations. This study tests that assumption by examining whether predictive accuracy and interpretability diverge in over-parameterised settings. To separate architectural effects from training variability, the experiments control model depth and weight initialisation to identify factors associated with explanation quality. We evaluate whether increased depth improves the semantic alignment of saliency maps or instead increases gradient noise that reduces their interpretability.

Understanding how architectural choices affect both predictive performance and explanation quality is therefore important. This study examines the relationship between model depth and interpretability in computer vision to test whether deeper networks produce higher-quality XAI explanations. The main contributions are as follows:
\begin{itemize}
    \item We examine the relationship between model complexity, characterised by the number of trainable parameters, and classification performance.
    \item We quantify how model scale and pretraining affect the quality of XAI explanations. We train eleven deep learning models on three datasets and generate explanations using five XAI methods.
    \item We evaluate the spatial precision of model predictions with respect to ground truth masks and report this alignment using quantitative measures.
    \item We introduce a new XAI metric, \emph{Dual-Polarity Precision} (DPP), which addresses the limitations of magnitude-only evaluation by explicitly measuring the spatial precision of both positive supporting evidence and negative counter-evidence, ensuring that explanations are penalized for contradictory or noisy background attributions.
\end{itemize}

\section*{Methods}
We evaluate nine ResNet and DenseNet models, together with two Vision Transformers (ViT-B/16 and ViT-L/16)~\cite{huang2017densely,dosovitskiy2020image}. Seven models are trained from scratch, and four use pretrained weights, including ResNet-50 pretrained on ImageNet-1k, DenseNet-121 pretrained on CheXpert, and ViT-B/16 and ViT-L/16 pretrained on ImageNet-21k~\cite{rajpurkar2017chexnet,irvin2019chexpert}. In total, the study includes 11 vision models that differ in architecture and parameter count.

Five XAI methods are used to obtain complementary attribution views of model behaviour. The methods are Saliency Maps~\cite{simonyan_deep_2014}, GradientSHAP~\cite{lundberg_unified_2017}, Integrated Gradients~\cite{sundararajan_axiomatic_2017}, Feature Permutation~\cite{breiman2001random,fisher2019all}, and Grad-CAM~\cite{selvaraju2017grad}. Explanations are evaluated with two quantitative metrics. Relevance Rank Accuracy~\cite{arras_ground_2022} measures agreement between attribution rankings and ground-truth masks, and the proposed Dual-Polarity Precision measures the fraction of positive pixel attributions that fall inside the ground truth mask and the fraction of negative attributions outside the class-mask. Figure~\ref{fig:overview} summarises the experimental workflow.

\subsection*{Datasets and Preprocessing}
\begin{figure*}[ht!]
    \centering
    \includegraphics[width=0.7\textwidth]{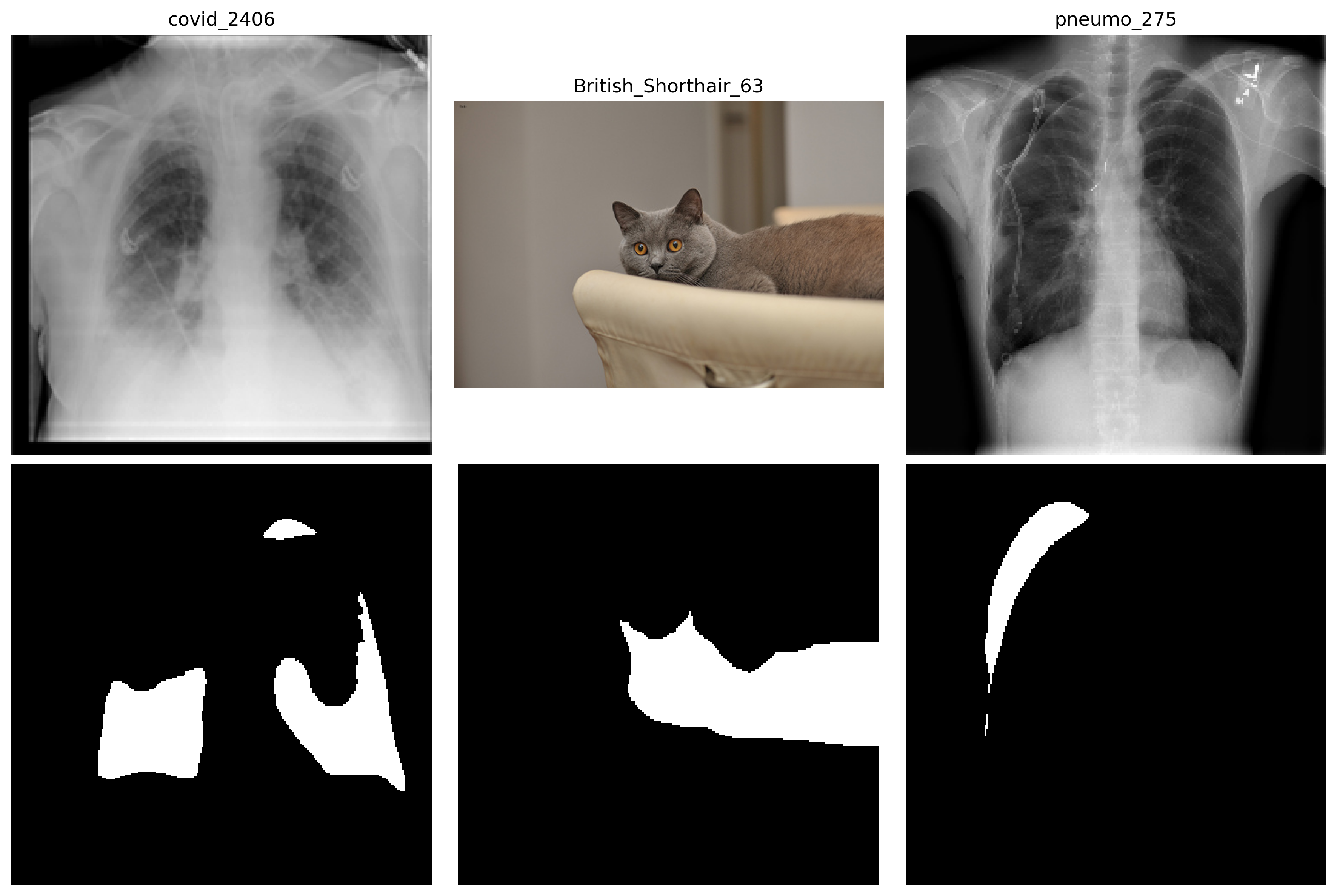}
    \caption{Examples of images from the datasets with corresponding ground-truth binary masks. Masks show the specific segments of interest (disease or an animal) used to evaluate the quality of the XAI explanations.}
    \label{fig:img_examples}
\end{figure*}

To assess model interpretability across different visual recognition settings, we use three datasets: COVID-QU-Ex\footnote{\url{https://www.kaggle.com/datasets/anasmohammedtahir/covidqu}},~\cite{thair_COVID-QU-Ex}, the Oxford-IIIT Pet dataset\footnote{\url{https://www.robots.ox.ac.uk/~vgg/data/pets/}}, and the Chest X-ray Pneumothorax dataset\footnote{\url{https://www.kaggle.com/datasets/vbookshelf/pneumothorax-chest-xray-images-and-masks}}. The datasets differ in size and task difficulty. Each provides binary ground truth masks that mark regions of interest, including infection areas annotated by clinicians in the medical datasets and object locations in the animal dataset. These masks enable quantitative comparison of generated explanations against annotated regions. Table~\ref{tab:images_masks} summarises the dataset characteristics, and Figure~\ref{fig:img_examples} shows example images with their corresponding masks.

\begin{table}[ht!]
    \centering
\caption{Image and ground-truth masks statistics per dataset.}
\label{tab:images_masks}

\begin{tabular}{lcccc}
    \toprule
    \multirow{2}{*}{Dataset} & 
    Number of &
    Number of &
    Average Mask &
    Average Mask/Image \\
     
    & Images
    & GT Masks 
    & Area (pixels)
    & Ratio (\%) \\
    \midrule
    COVID-Qu-Ex & $4{,}369$ & $3{,}812$ & $9{,}076_{\pm 5{,}147}$ & $18.1_{\pm 10.3}$ \\
    Oxford-IIIT Pet & $7{,}390$ & $7{,}349$ & $14{,}885_{\pm 8{,}238}$ & $29.7_{\pm 16.4}$ \\
    Chest X Pneumothorax & $12{,}047$ & $2{,}669$ & $709_{\pm 798}$ & $1.4_{\pm 1.6}$ \\
    \bottomrule
\end{tabular}
\end{table}

Images in each dataset are randomly split into training (70\%), validation (20\%), and test (10\%) sets. All inputs are resized to $224 \times 224$ pixels, converted to single-channel greyscale, and normalised using a mean and standard deviation of 0.5. Data augmentation is applied only to the training set to support generalisation, including random horizontal flips with probability 0.5 and random rotations up to 15 degrees. Validation and test images undergo only resizing and normalisation to keep evaluation conditions consistent with the original data distribution.

\subsection*{Models}
We study nine convolutional models and two Vision Transformers to examine how model complexity relates to the quality of XAI explanations. Seven CNNs are trained from scratch, including ResNet-18, ResNet-34, ResNet-50, and ResNet-101~\cite{he_deep_2015}, as well as DenseNet-121, DenseNet-169, and DenseNet-201~\cite{densely}. Training these models under a shared setup supports comparisons across depths within each architecture family. 

Four additional models use pretrained weights to represent common transfer-learning settings. These include two CNN pretrained models, namely ResNet-50 pretrained on ImageNet-1K\_V2 dataset~\cite{imagenet} containing over a million images, and a DenseNet-121 pretrained on CheXpert dataset~\cite{chexpert} containing 224,316 chest radiographs. We also include two Vision Transformers, ViT-B/16 (224, IN21k) and ViT-L/16 (224, IN21k), pretrained on more than 14 million images (ImageNet-21k)~\cite{dosovitskiy2020image}. Overall, the experiments cover 11 vision models that differ in architecture and parameter count. Table~\ref{tab:parameters} reports the number of trainable parameters for each model.

\subsection*{Model Training Setup}
We treat each task as binary classification and train with a class-weighted binary cross-entropy loss to address class imbalance. For the seven models trained from scratch, we use AdamW with learning rate 0.001, $\beta_1=0.9$, $\beta_2=0.999$, and weight decay 0.01. We apply a \texttt{ReduceLROnPlateau} scheduler that reduces the learning rate by a factor of 0.5 when the validation loss does not improve for five epochs. Training uses a batch size of 32 for up to 100 epochs, with early stopping based on class-weighted validation loss and a patience of 12 epochs. The 100-epoch setting serves as a shared upper budget across architectures, while the effective training duration is determined adaptively by validation performance.

Pretrained models are fine-tuned in two stages. First, we freeze the feature extractor and train only the classification head for up to 100 epochs, using the same scheduler and early-stopping settings. Second, we unfreeze all layers and continue end-to-end training with the same scheduler while reducing the learning rate to $1\times 10^{-5}$.

For reproducibility, we run each experiment with three different random seeds. For each seed, we train the models and perform the XAI analysis.

\begin{table}[t!]
    \centering
\caption{Total number of parameters (in millions) for each architecture.}
\label{tab:parameters}
\begin{tabular}{lc}
    \toprule
    Model & Number of Parameters (M) \\
    \midrule
    ResNet-18 & 11.2 \\
    ResNet-34 & 21.3 \\
    ResNet-50 & 23.5 \\
    ResNet-101 & 42.5 \\
    DenseNet-121 & 6.9 \\
    DenseNet-169 & 12.5 \\
    DenseNet-201 & 18.1 \\
    ViT-B/16 (224) & 85.5 \\
    ViT-L/16 (224) & 302.8 \\
    \bottomrule
\end{tabular}
\end{table}

\subsection*{XAI Methods}
We apply five explanation methods that cover gradient-based and perturbation-based approaches. Saliency maps compute pixel-level importance from the gradient of the class score with respect to the input image~\cite{simonyan_deep_2014}. Integrated Gradients addresses gradient saturation by accumulating gradients along a path from a baseline input to the observed image~\cite{sundararajan_axiomatic_2017}. GradientSHAP approximates Shapley-style attributions by combining stochastic baselines with gradient information for efficient estimation~\cite{lundberg_unified_2017}.

We also use Grad-CAM, which produces a coarse localisation map by weighting feature maps in the last convolutional layer according to their contribution to the predicted class~\cite{selvaraju2017grad}. Finally, Feature Permutation method estimates importance by shuffling selected input features and measuring the resulting change in the model output~\cite{breiman2001random,fisher2019all}.

\subsubsection*{Saliency Maps}
Saliency Maps represent a foundational methodology in XAI. This technique visualises the significance of specific pixels regarding the model predictions. Deep neural networks, due to their complexity and nonlinearity, are not inherently interpretable, unlike simpler models such as linear and logistic regression. Saliency maps address this difficulty by highlighting image regions that exhibit strong correlations to specific classes. Simonyan et al. (2014)~\cite{simonyan_deep_2014} introduced a technique based on gradients to compute saliency maps. Their method ranks pixels according to their impact on the class score. It approximates the complex function linearly using a Taylor expansion. The procedure involves determining the derivative via backpropagation. The final map displays the magnitude of the gradients.

\subsubsection*{Integrated Gradients}
Integrated Gradients addresses the axiomatic deficiencies found in earlier attribution techniques. Sundararajan et al. (2017)~\cite{sundararajan_axiomatic_2017} introduced this method to satisfy two fundamental properties known as sensitivity and implementation invariance. Sensitivity ensures that any feature responsible for a prediction change receives a non-zero attribution. Implementation invariance guarantees that functionally equivalent networks produce identical explanations regardless of their internal structure. The method defines feature importance by calculating the path integral of the gradients. It traverses a linear path from a baseline input to the actual input. In image classification tasks, the baseline is typically a black image. The integral is often approximated using a Riemann sum. This approach yields a complete attribution where the sum of importance scores equals the difference between the model output for the input and the baseline.

\subsubsection*{GradientSHAP}
GradientSHAP links gradient-based attribution to Shapley values. Shapley values come from cooperative game theory and assign each player an allocation based on its average marginal contribution across coalitions. Lundberg and Lee (2017)~\cite{lundberg_unified_2017} adapt this idea to machine learning in the SHAP framework, which assigns a contribution score to each feature while satisfying properties such as local accuracy and consistency.

GradientSHAP provides a practical approximation for deep networks by estimating expected gradients along interpolations between a baseline and the input. The method samples points on this path and adds Gaussian noise to the input, then computes gradients of the model output with respect to the perturbed samples. Attributions are obtained by multiplying these gradients by the input–baseline difference and averaging across samples. The approximation relies on assumptions commonly used in SHAP-style methods, including feature independence and approximately linear behaviour between the baseline and the input.

\begin{figure*}[t!]
\centering
\begin{subfigure}[t]{\textwidth}
  \centering
  \subcaption{COVID-19 X-ray image example.}
  \includegraphics[width=0.8\textwidth]{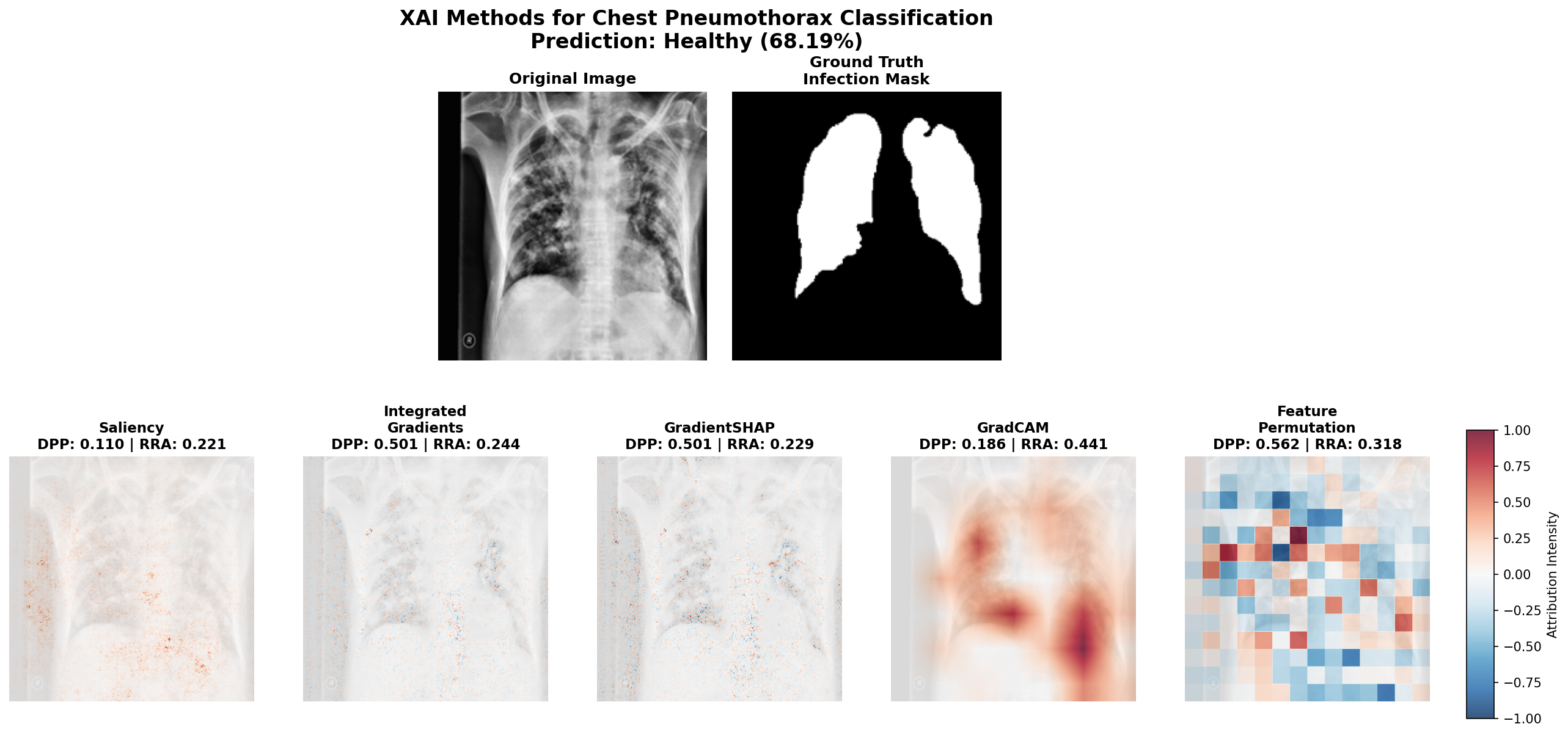}
\label{fig:xai_salency_example_covid}
\end{subfigure}

\vspace{0.6em}
\begin{subfigure}[t]{\textwidth}
  \centering
  \subcaption{Pneumothorax X-ray image example.}
  \includegraphics[width=0.8\textwidth]{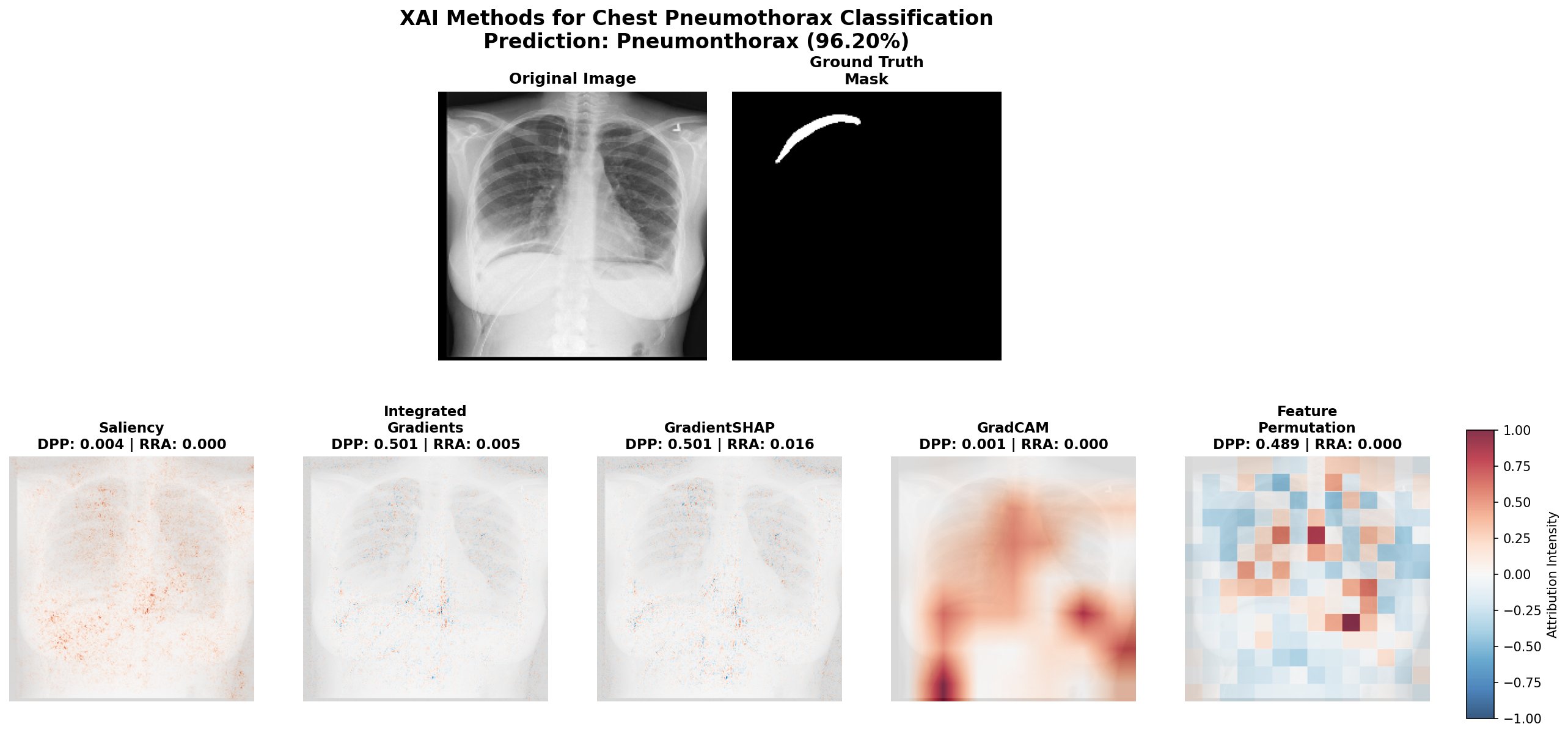}
\label{fig:xai_salency_example_pneumo}
\end{subfigure}
\caption{\textbf{Qualitative saliency examples on chest radiographs.}
(a) COVID-QU-Ex example and (b) Pneumothorax example. For each case, the top row shows the input X-ray, the corresponding expert-annotated binary ground-truth mask, and the model prediction with confidence as shown in the panel header. The bottom row shows attribution heatmaps generated by Saliency, Integrated Gradients, GradientSHAP, Grad-CAM, and Feature Permutation, with the localisation scores (RRA and DPP) reported above each map. Heatmaps show the importance regions, where warmer colours indicate higher attribution and cooler colours indicate lower attribution values. Grad-CAM is upsampled from the final convolutional feature-map resolution, and Feature Permutation is computed over $16\times16$ patches and therefore shows blockwise structure.}
\label{fig:xai_salency_example}
\end{figure*}

\subsubsection*{Grad-CAM}
Gradient-weighted Class Activation Mapping (Grad-CAM)~\cite{selvaraju2017grad} uses the spatial information retained in the final convolutional layer of a neural network to produce visual explanations. This method computes the gradients of the target score with respect to the feature maps of the last convolutional layer. These gradients are globally averaged to obtain importance weights for each channel. The algorithm calculates a weighted combination of the forward activation maps using these importance weights. A Rectified Linear Unit function is applied to the result to isolate features that have a positive influence on the class of interest. This ensures that the heatmap highlights pixels that contribute to the prediction rather than those that suppress it. The resulting coarse localisation map is upsampled to the resolution of the input image. This process reveals the specific regions responsible for the decision of the model without requiring architectural changes or retraining. For Vision Transformers, Grad-CAM was adapted by targeting the layer normalization after the last transformer encoder block (\texttt{encoder.layer[-1].layernorm\_after}) rather than a convolutional layer. This follows common practice when applying CAM-style methods to ViTs, where a normalization layer in the last block is used as the attribution layer and token activations are reshaped into a spatial map~\cite{chefer2021transformer}. We implemented this using Captum's \texttt{LayerGradCam}~\cite{kokhlikyan2020captum}, and the resulting heatmap was upsampled to the input resolution using bilinear interpolation.

\subsubsection*{Feature Permutation}
Feature Permutation, also called permutation feature importance, measures how much a trained model relies on an input feature by comparing model performance before and after that feature is permuted on an evaluation set~\cite{breiman2001random,fisher2019all}. The baseline metric is computed on the original inputs. For each feature group $j$, the method then creates a perturbed batch by shuffling the values of group $j$ across samples while keeping all other input values fixed. The model is evaluated on this perturbed batch, and the importance of group $j$ is defined as the change in the metric relative to the baseline. The permutation is repeated multiple times and the results are averaged to reduce sampling variance.

For images, we define feature groups as non-overlapping $16 \times 16$ pixel patches on the resized $224 \times 224$ inputs, giving a $14 \times 14$ grid of 196 patches. Group membership is encoded by an integer-valued mask $M \in \{0,\dots,195\}^{1 \times 224 \times 224}$ where all pixels in the same patch share the same index. During permutation, all pixels that share an index are shuffled together across the batch, so the same sample-level permutation is applied to every pixel in the patch. The resulting patch score is then assigned to all pixels in that patch to obtain a patch-level importance map.

\subsection*{Quantifying the Quality of XAI Explanations}
In this study, we define explanation quality as localisation agreement between an attribution heatmap and an expert-annotated binary region-of-interest mask. The evaluation therefore measures how well the explanation concentrates relevance inside the annotated region, rather than broader notions of human usefulness~\cite{arras_ground_2022,hedstrom2023quantus}.

We compare each heatmap with its ground truth mask using Relevance Rank Accuracy, which reports the fraction of the highest-ranked attribution locations that lie inside the mask, and \emph{Dual-Polarity Precision} (DPP), which reports a balanced measure of spatial precision by evaluating the localization of both positive supporting evidence and negative counter-evidence.

\subsubsection*{Relevance Rank Accuracy}
Relevance Rank Accuracy (RRA)~\cite{arras_ground_2022} evaluates the spatial alignment between the most salient relevance scores and the ground truth. Let \( GT \) denote the set of pixel coordinates within the ground truth mask, and let \( K = |GT| \) represent the total number of pixels in this mask. To compute the RRA, we extract the top \( K \) pixels with the highest relevance values from the input image. The accuracy is then defined as the intersection over \( K \), measuring the proportion of these highest-ranking pixels that fall within the ground truth mask.

Formally, we define the set of the top \( K \) pixel coordinates as:

\begin{equation}
P_{K} = \{ p_1, p_2, \ldots, p_K \}
\end{equation}

such that their corresponding relevance values \( R \) satisfy \( R_{p_1} \geq R_{p_2} \geq \ldots \geq R_{p_K} \), and for any pixel \( q \notin P_{K} \), it holds that \( R_{p_K} \geq R_{q} \). Consequently, the Relevance Rank Accuracy is calculated as:

\begin{equation}
\text{RRA} = \frac{|P_{K} \cap GT|}{|GT|}
\end{equation}

where $|\text{GT}|$ is the total number of pixels in the ground truth mask.

\subsubsection*{Dual-Polarity Precision}
Arras et al.\ evaluate localisation under the assumption that the major part of relevance should lie within the ground truth mask, either in terms of relevance mass or relevance ranking, and their metrics do not require binarising the heatmap~\cite{arras_ground_2022}. In practice, mass-based evaluation is applied to a pooled heatmap, and common pooling choices map signed attributions to non-negative values, for example by squaring or rectifying per-channel relevances before aggregation~\cite{arras_ground_2022}. This pooling can remove sign information, which means that negative attributions contribute by magnitude rather than acting as counter-evidence~\cite{balogh2023evaluation}.

In our setting, the expert mask specifies regions expected to contain evidence for the target label. We therefore report a sign-aware metric that evaluates the spatial precision of both (i) positive attributions that indicate supporting evidence and (ii) negative attributions that indicate counter-evidence. We define the positive and negative parts of the attribution map as $R^{+}(p)=\max(R(p),0)$ and $R^{-}(p)=\max(-R(p),0)$. We then compute
\begin{align}
TPA &= \sum_{p \in GT} R^{+}(p), &
FPA &= \sum_{p \in \Omega \setminus GT} R^{+}(p), \\
TNA &= \sum_{p \in \Omega \setminus GT} R^{-}(p), &
FNA &= \sum_{p \in GT} R^{-}(p),
\end{align}
where $GT$ is the set of pixel locations in the ground truth mask and $\Omega$ denotes all pixel locations. The precision of the positive signal is
\begin{equation}
P_{\text{pos}} = \frac{TPA}{TPA + FPA + \epsilon},
\end{equation}
which corresponds to a mass-precision variant of relevance mass accuracy as implemented in common XAI evaluation toolkits~\cite{hedstrom2023quantus}. The precision of the negative signal is
\begin{equation}
P_{\text{neg}} = \frac{TNA}{TNA + FNA + \epsilon},
\end{equation}
which measures whether negatively signed attribution mass is concentrated outside the annotated region. Finally, we define \emph{Dual-Polarity Precision} as a macro-average of the two terms,
\begin{equation}
\mathrm{DPP} = \frac{P_{\text{pos}} + P_{\text{neg}}}{2}.
\end{equation}
DPP takes high values when supporting attribution mass concentrates within the expert mask and negative attribution mass lies primarily outside the mask. We use $\epsilon$ to avoid numerical instability when the total positive or negative mass is close to zero.

\subsubsection*{Complementary use of RRA and DPP}
While RRA and DPP may correlate, they measure distinct aspects of localisation performance. RRA relies on the ranking of attribution values within the top-$K$ pixels. This metric remains robust to sparse but high-magnitude noise outside the target region. DPP operates as a mass-based metric that accounts for the total magnitude and sign of the attribution signal. Unlike RRA, which focuses on the highest-ranking features, DPP penalizes accumulated attribution mass that contradicts the ground truth. This penalty applies to positive mass in the background and negative mass within the region of interest. A method may therefore achieve a high RRA score by correctly positioning peak pixels but receive a lower DPP score if it generates considerable background noise. 

The interpretation of the DPP score depends on the polarity of the attribution. A score of 0.5 represents a baseline of random attribution allocation for methods that produce signed attributions. For methods limited to positive attributions the baseline is 0 and the highest possible alignment score is 0.5. We use both metrics to distinguish between a method that achieves local pixel-wise accuracy at its peaks and another that provides an explanation across the entire image.

\section*{Results}
\label{sec:results}

\subsection*{Models' Performance} Table~\ref{tab:model_performance} summarizes the classification performance results. All evaluated architectures achieve high to near-perfect performance on the COVID-QU-Ex and Oxford-IIIT Pet datasets. The CheXpert model yields the lowest AUC ($0.85$) on the Oxford-IIIT Pet dataset. This outcome supports the expectation that medical pretraining generalizes poorly to natural images. Smaller models trained from scratch, such as ResNet-18 and DenseNet-121, match or exceed the performance of larger architectures that contain millions more parameters.

The Chest X Pneumothorax dataset presents the most challenging classification task and results in lower metrics across all models. AUC scores range from $0.77$ for the standard ResNet-50 to $0.87$ for the pretrained ResNet-50 and the scratch-trained DenseNet-201. Consistent with results from the other datasets, smaller architectures such as ResNet-18 and DenseNet-121 achieve performance levels comparable to or better than their deeper counterparts. We verify the training dynamics for all 99 experimental runs to ensure the validity of the subsequent XAI evaluation.

We monitor training and validation loss curves to confirm convergence and prevent overfitting. The application of early stopping ensures that all models are evaluated at their point of optimal generalization. Only the CheXpert-pretrained DenseNet-121 fails to converge across three different seed runs for the COVID-QU-Ex and Oxford-IIIT Pet datasets. In contrast, this model converges successfully on the Chest X Pneumothorax dataset and achieves the second-best result.

\begin{table}[t!]
\centering
\small
\caption{Performance on the test set (means $\pm$ standard deviations) averaged across three randomly drawn seeds. Loss values reflect the final validation loss at the best model checkpoint. Training and inference times are reported in seconds. Bolded values indicate the best result per metric, whereas underlined values indicate the second-best. Note: the \textsuperscript{$\dagger$} denotes models for which none of the three seeded runs satisfied the convergence criterion within the 100-epoch budget.}
\label{tab:model_performance}
\resizebox{\textwidth}{!}{%
\begin{tabular}{llcccc}
\toprule
\makecell[l]{Dataset\\(\# images)} & Model & AUC $\uparrow$ & Loss $\downarrow$ &
\makecell{Train\\Time (s) $\downarrow$} &
\makecell{Inference\\Time (s) $\downarrow$} \\
\midrule
\multirow{11}{*}{\shortstack[l]{\textit{COVID-QU-Ex}\\ (4,369)}}
& ResNet-18             & \underline{$0.99_{\pm 0.00}$} & $0.08_{\pm 0.01}$ & $\mathbf{673}_{\pm 82}$  & \underline{$1.22_{\pm 0.01}$} \\
& ResNet-34             & \underline{$0.99_{\pm 0.00}$} & $0.10_{\pm 0.01}$ & \underline{$732_{\pm 173}$} & $1.29_{\pm 0.04}$ \\
& ResNet-50             & \underline{$0.99_{\pm 0.01}$} & $0.07_{\pm 0.02}$ & $797_{\pm 157}$ & $1.24_{\pm 0.06}$ \\
& ResNet-50 (IN-1k)    & $\mathbf{1.00}_{\pm 0.00}$ & $\mathbf{0.05}_{\pm 0.02}$ & $1227_{\pm 218}$ & $\mathbf{1.15}_{\pm 0.02}$ \\
& ResNet-101            & $\mathbf{1.00}_{\pm 0.00}$ & $\mathbf{0.05}_{\pm 0.01}$ & $1057_{\pm 173}$ & $1.35_{\pm 0.05}$ \\
& DenseNet-121          & $\mathbf{1.00}_{\pm 0.00}$ & $0.07_{\pm 0.02}$ & $1219_{\pm 259}$ & $1.41_{\pm 0.13}$ \\
& DenseNet-121 (CheXpert) & $0.98_{\pm 0.01}$ & $0.16_{\pm 0.06}$ & $1468_{\pm 135}$\textsuperscript{$\dagger$} & $1.50_{\pm 0.06}$ \\
& DenseNet-169          & $\mathbf{1.00}_{\pm 0.00}$ & $0.07_{\pm 0.02}$ & $1245_{\pm 282}$ & $1.41_{\pm 0.08}$ \\
& DenseNet-201          & \underline{$0.99_{\pm 0.00}$} & $0.08_{\pm 0.00}$ & $1312_{\pm 115}$ & $1.51_{\pm 0.10}$ \\
& ViT-B/16 (224, IN-21k) & $\mathbf{1.00}_{\pm 0.00}$ & \underline{$0.06_{\pm 0.01}$} & $2713_{\pm 126}$ & $2.04_{\pm 0.09}$ \\
& ViT-L/16 (224, IN-21k) & \underline{$0.99_{\pm 0.00}$} & $0.11_{\pm 0.03}$ & $6111_{\pm 744}$ & $4.59_{\pm 0.06}$ \\

\midrule
\multirow{11}{*}{\shortstack[l]{\textit{Oxford-IIIT Pet}\\ (7,390)}}
& ResNet-18             & $0.98_{\pm 0.01}$ & $0.19_{\pm 0.01}$ & $\mathbf{2761}_{\pm 189}$   & $7.12_{\pm 0.61}$ \\
& ResNet-34             & \underline{$0.99_{\pm 0.00}$} & $0.16_{\pm 0.01}$ & $3684_{\pm 165}$   & $\mathbf{6.92}_{\pm 0.63}$ \\
& ResNet-50             & \underline{$0.99_{\pm 0.00}$} & $0.17_{\pm 0.02}$ & \underline{$3551_{\pm 465}$}   & $7.42_{\pm 0.45}$ \\
& ResNet-50 (IN-1k)    & $\mathbf{1.00}_{\pm 0.00}$ & $\mathbf{0.01}_{\pm 0.00}$ & $4032_{\pm 577}$   & $7.31_{\pm 0.55}$ \\
& ResNet-101            & \underline{$0.99_{\pm 0.01}$} & $0.15_{\pm 0.01}$ & $3857_{\pm 697}$   & $7.51_{\pm 0.28}$ \\
& DenseNet-121          & \underline{$0.99_{\pm 0.00}$} & $0.16_{\pm 0.03}$ & $4039_{\pm 170}$   & \underline{$7.04_{\pm 0.28}$} \\
& DenseNet-121 (CheXpert) & $0.85_{\pm 0.01}$ & $0.49_{\pm 0.01}$ & $6379_{\pm 65}$\textsuperscript{$\dagger$} & $7.58_{\pm 0.63}$ \\
& DenseNet-169          & \underline{$0.99_{\pm 0.00}$} & $0.14_{\pm 0.03}$ & $4024_{\pm 1041}$  & $7.45_{\pm 0.17}$ \\
& DenseNet-201          & \underline{$0.99_{\pm 0.00}$} & $0.15_{\pm 0.02}$ & $4196_{\pm 295}$   & $7.70_{\pm 0.11}$ \\
& ViT-B/16 (224, IN-21k) & $\mathbf{1.00}_{\pm 0.00}$ & \underline{$0.02_{\pm 0.00}$} & $4828_{\pm 1068}$  & $8.65_{\pm 0.21}$ \\
& ViT-L/16 (224, IN-21k) & $\mathbf{1.00}_{\pm 0.00}$ & $0.01_{\pm 0.00}$ & $10188_{\pm 1325}$ & $12.04_{\pm 0.28}$ \\

\midrule
\multirow{11}{*}{\shortstack[l]{\textit{Chest X Pneumothorax}\\ (12,047)}}
& ResNet-18                 & \underline{$0.86_{\pm 0.01}$} & \underline{$0.47_{\pm 0.01}$} & \underline{$1544_{\pm 10}$}    & $\mathbf{3.14}_{\pm 0.02}$ \\
& ResNet-34                 & \underline{$0.86_{\pm 0.01}$} & $0.48_{\pm 0.01}$ & $1981_{\pm 110}$   & \underline{$3.21_{\pm 0.04}$} \\
& ResNet-50                 & $0.77_{\pm 0.07}$ & $0.57_{\pm 0.07}$ & $\mathbf{1034}_{\pm 491}$   & $3.33_{\pm 0.04}$ \\
& ResNet-50 (IN-1k)          & $\mathbf{0.87}_{\pm 0.00}$ & $\mathbf{0.46}_{\pm 0.00}$ & $2056_{\pm 560}$   & $3.42_{\pm 0.12}$ \\
& ResNet-101                & \underline{$0.86_{\pm 0.00}$} & $0.48_{\pm 0.00}$ & $2846_{\pm 132}$   & $3.80_{\pm 0.11}$ \\
& DenseNet-121              & \underline{$0.86_{\pm 0.01}$} & $\mathbf{0.46}_{\pm 0.01}$ & $2712_{\pm 232}$   & $3.81_{\pm 0.10}$ \\
& DenseNet-121 (CheXpert)   & \underline{$0.86_{\pm 0.01}$} & \underline{$0.47_{\pm 0.01}$} & $4034_{\pm 82}$    & $3.78_{\pm 0.17}$ \\
& DenseNet-169              & \underline{$0.86_{\pm 0.01}$} & $\mathbf{0.46}_{\pm 0.00}$ & $2814_{\pm 160}$   & $4.05_{\pm 0.06}$ \\
& DenseNet-201              & $\mathbf{0.87}_{\pm 0.01}$ & $\mathbf{0.46}_{\pm 0.01}$ & $3771_{\pm 114}$   & $4.43_{\pm 0.18}$ \\
& ViT-B/16 (224, IN-21k)     & $0.83_{\pm 0.02}$ & $0.51_{\pm 0.03}$ & $5798_{\pm 911}$   & $5.59_{\pm 0.14}$ \\
& ViT-L/16 (224, IN-21k)     & $0.80_{\pm 0.02}$ & $0.55_{\pm 0.03}$ & $15776_{\pm 999}$  & $11.76_{\pm 0.06}$ \\

\bottomrule
\end{tabular}
}
\end{table}

\begin{table}[t!]
\centering
\caption{Evaluation of explanations obtained from five XAI methods applied on 200 randomly drawn images from the test sets per seed (mean $\pm$ standard deviation averaged across 600 images sampled over three seeds). Bolded values indicate the highest scores per category, whereas underlined values are the second-best results. Note: \textsuperscript{$\dagger$} indicates model that did not converge within the training budget of 100 epochs across all seeds.}
\label{tab:xai_results}
\resizebox{\textwidth}{!}{%
\begin{tabular}{lcccccccccc}
\toprule
\multirow{2.5}{*}{Model} &
\multicolumn{5}{c}{Relevance Rank Accuracy $\uparrow$} &
\multicolumn{5}{c}{Dual-Polarity Precision $\uparrow$} \\
\cmidrule(lr){2-6} \cmidrule(lr){7-11}
& FeatPerm & Grad-CAM & GradSHAP & IntGrad & Saliency & FeatPerm & Grad-CAM & GradSHAP & IntGrad & Saliency \\
\midrule

\multicolumn{11}{c}{\textit{COVID-Qu-Ex}} \\
ResNet-18
& $0.345_{\pm 0.00}$ & $0.258_{\pm 0.01}$ & $\underline{0.228_{\pm 0.01}}$ & $\underline{0.228_{\pm 0.01}}$ & $0.283_{\pm 0.01}$ & $0.570_{\pm 0.01}$ & $0.122_{\pm 0.01}$ & $\underline{0.501_{\pm 0.00}}$ & $0.500_{\pm 0.00}$ & $\underline{0.499_{\pm 0.00}}$ \\
ResNet-34
& $\underline{0.347_{\pm 0.00}}$ & $0.219_{\pm 0.03}$ & $\bm{0.233_{\pm 0.01}}$ & $\bm{0.235_{\pm 0.01}}$ & $0.290_{\pm 0.01}$ & $0.576_{\pm 0.02}$ & $0.110_{\pm 0.01}$ & $0.497_{\pm 0.00}$ & $0.498_{\pm 0.00}$ & $\underline{0.499_{\pm 0.00}}$ \\
ResNet-50
& $0.279_{\pm 0.03}$ & $0.229_{\pm 0.01}$ & $0.216_{\pm 0.01}$ & $0.214_{\pm 0.01}$ & $0.288_{\pm 0.02}$ & $0.534_{\pm 0.02}$ & $0.110_{\pm 0.00}$ & $0.499_{\pm 0.00}$ & $0.498_{\pm 0.00}$ & $0.497_{\pm 0.00}$ \\
ResNet-50 (IN-1k)
& $0.292_{\pm 0.01}$ & $0.266_{\pm 0.02}$ & $0.198_{\pm 0.00}$ & $0.201_{\pm 0.00}$ & $0.247_{\pm 0.01}$ & $0.541_{\pm 0.00}$ & $0.120_{\pm 0.01}$ & $0.499_{\pm 0.00}$ & $0.499_{\pm 0.00}$ & $\bm{0.500_{\pm 0.00}}$ \\
ResNet-101
& $0.301_{\pm 0.01}$ & $0.253_{\pm 0.01}$ & $0.226_{\pm 0.01}$ & $0.226_{\pm 0.01}$ & $0.286_{\pm 0.00}$ & $0.542_{\pm 0.02}$ & $0.114_{\pm 0.01}$ & $\bm{0.502_{\pm 0.00}}$ & $\bm{0.503_{\pm 0.00}}$ & $0.497_{\pm 0.00}$ \\
DenseNet-121
& $0.336_{\pm 0.06}$ & $0.221_{\pm 0.05}$ & $0.209_{\pm 0.01}$ & $0.211_{\pm 0.01}$ & $0.283_{\pm 0.01}$ & $0.602_{\pm 0.04}$ & $0.107_{\pm 0.01}$ & $0.497_{\pm 0.00}$ & $0.497_{\pm 0.00}$ & $0.497_{\pm 0.00}$ \\
DenseNet-121 (CheXpert)\textsuperscript{$\dagger$}
& $0.258_{\pm 0.01}$ & $\bm{0.349_{\pm 0.02}}$ & $0.225_{\pm 0.00}$ & $\underline{0.228_{\pm 0.00}}$ & $\underline{0.294_{\pm 0.00}}$ & $0.478_{\pm 0.01}$ & $\bm{0.149_{\pm 0.00}}$ & $0.500_{\pm 0.00}$ & $0.500_{\pm 0.00}$ & $0.497_{\pm 0.00}$ \\
DenseNet-169
& $\bm{0.377_{\pm 0.03}}$ & $0.285_{\pm 0.04}$ & $0.223_{\pm 0.01}$ & $0.226_{\pm 0.01}$ & $\bm{0.296_{\pm 0.01}}$ & $\bm{0.614_{\pm 0.01}}$ & $0.123_{\pm 0.01}$ & $0.500_{\pm 0.00}$ & $0.499_{\pm 0.00}$ & $0.495_{\pm 0.00}$ \\
DenseNet-201
& $0.344_{\pm 0.03}$ & $\underline{0.323_{\pm 0.04}}$ & $0.206_{\pm 0.01}$ & $0.210_{\pm 0.01}$ & $0.265_{\pm 0.01}$ & $\underline{0.603_{\pm 0.02}}$ & $\underline{0.134_{\pm 0.01}}$ & $0.500_{\pm 0.00}$ & $\underline{0.501_{\pm 0.00}}$ & $0.497_{\pm 0.00}$ \\
ViT-B/16 (224, IN-21k)
& $0.302_{\pm 0.02}$ & $0.195_{\pm 0.00}$ & $0.213_{\pm 0.01}$ & $0.211_{\pm 0.01}$ & $0.293_{\pm 0.01}$ & $0.510_{\pm 0.00}$ & $0.097_{\pm 0.00}$ & $0.495_{\pm 0.00}$ & $0.495_{\pm 0.00}$ & $0.497_{\pm 0.00}$ \\
ViT-L/16 (224, IN-21k)
& $0.290_{\pm 0.00}$ & $0.203_{\pm 0.01}$ & $0.214_{\pm 0.01}$ & $0.214_{\pm 0.02}$ & $0.278_{\pm 0.00}$ & $0.513_{\pm 0.00}$ & $0.101_{\pm 0.00}$ & $0.499_{\pm 0.00}$ & $0.498_{\pm 0.00}$ & $\underline{0.499_{\pm 0.00}}$ \\
\midrule

\multicolumn{11}{c}{\textit{Oxford-IIIT Pet}} \\
ResNet-18
& $0.313_{\pm 0.02}$ & $0.368_{\pm 0.03}$ & $0.345_{\pm 0.01}$ & $0.347_{\pm 0.01}$ & $0.337_{\pm 0.01}$
& $0.490_{\pm 0.01}$ & $0.181_{\pm 0.02}$ & $\underline{0.502_{\pm 0.00}}$ & $0.502_{\pm 0.00}$ & $\bm{0.500_{\pm 0.00}}$ \\
ResNet-34
& $0.309_{\pm 0.01}$ & $0.360_{\pm 0.02}$ & $0.346_{\pm 0.01}$ & $0.347_{\pm 0.02}$ & $0.341_{\pm 0.01}$
& $0.486_{\pm 0.01}$ & $0.174_{\pm 0.01}$ & $\underline{0.502_{\pm 0.00}}$ & $0.501_{\pm 0.00}$ & $\bm{0.500_{\pm 0.00}}$ \\
ResNet-50
& $0.328_{\pm 0.02}$ & $0.373_{\pm 0.01}$ & $0.336_{\pm 0.01}$ & $0.339_{\pm 0.01}$ & $0.331_{\pm 0.01}$
& $\bm{0.509_{\pm 0.01}}$ & $0.181_{\pm 0.01}$ & $0.500_{\pm 0.00}$ & $0.500_{\pm 0.00}$ & $\underline{0.499_{\pm 0.00}}$ \\
ResNet-50 (IN-1k)
& $\bm{0.372_{\pm 0.02}}$ & $\bm{0.458_{\pm 0.04}}$ & $\bm{0.356_{\pm 0.01}}$ & $\bm{0.358_{\pm 0.01}}$ & $\bm{0.367_{\pm 0.01}}$
& $\underline{0.508_{\pm 0.01}}$ & $\bm{0.225_{\pm 0.02}}$ & $0.500_{\pm 0.00}$ & $0.500_{\pm 0.00}$ & $\bm{0.500_{\pm 0.00}}$ \\
ResNet-101
& $0.309_{\pm 0.01}$ & $\underline{0.411_{\pm 0.02}}$ & $0.349_{\pm 0.01}$ & $0.348_{\pm 0.01}$ & $0.345_{\pm 0.01}$
& $0.496_{\pm 0.01}$ & $\underline{0.197_{\pm 0.01}}$ & $\underline{0.502_{\pm 0.00}}$ & $0.502_{\pm 0.00}$ & $\bm{0.500_{\pm 0.00}}$ \\
DenseNet-121
& $0.293_{\pm 0.01}$ & $0.383_{\pm 0.04}$ & $0.354_{\pm 0.01}$ & $\underline{0.357_{\pm 0.01}}$ & $0.352_{\pm 0.01}$
& $0.453_{\pm 0.01}$ & $0.184_{\pm 0.01}$ & $\bm{0.503_{\pm 0.00}}$ & $\bm{0.504_{\pm 0.00}}$ & $\bm{0.500_{\pm 0.00}}$ \\
DenseNet-121 (CheXpert)\textsuperscript{$\dagger$}
& $0.317_{\pm 0.01}$ & $0.357_{\pm 0.03}$ & $0.329_{\pm 0.01}$ & $0.328_{\pm 0.01}$ & $0.328_{\pm 0.01}$
& $0.506_{\pm 0.01}$ & $0.173_{\pm 0.01}$ & $0.499_{\pm 0.00}$ & $0.500_{\pm 0.00}$ & $\underline{0.499_{\pm 0.00}}$ \\
DenseNet-169
& $0.324_{\pm 0.02}$ & $0.406_{\pm 0.03}$ & $0.354_{\pm 0.01}$ & $0.355_{\pm 0.01}$ & $0.351_{\pm 0.01}$
& $0.500_{\pm 0.01}$ & $0.191_{\pm 0.01}$ & $\bm{0.503_{\pm 0.00}}$ & $\underline{0.503_{\pm 0.00}}$ & $\bm{0.500_{\pm 0.00}}$ \\
DenseNet-201
& $0.317_{\pm 0.01}$ & $0.368_{\pm 0.02}$ & $\underline{0.355_{\pm 0.02}}$ & $0.356_{\pm 0.02}$ & $\underline{0.356_{\pm 0.01}}$
& $0.488_{\pm 0.02}$ & $0.178_{\pm 0.01}$ & $0.501_{\pm 0.00}$ & $0.502_{\pm 0.00}$ & $\bm{0.500_{\pm 0.00}}$ \\
ViT-B/16 (224, IN-21k)
& $\underline{0.338_{\pm 0.02}}$ & $0.313_{\pm 0.01}$ & $0.348_{\pm 0.01}$ & $0.350_{\pm 0.01}$ & $0.343_{\pm 0.02}$
& $0.500_{\pm 0.00}$ & $0.155_{\pm 0.01}$ & $0.501_{\pm 0.00}$ & $0.501_{\pm 0.00}$ & $\underline{0.499_{\pm 0.00}}$ \\
ViT-L/16 (224, IN-21k)
& $0.323_{\pm 0.01}$ & $0.315_{\pm 0.02}$ & $0.325_{\pm 0.01}$ & $0.328_{\pm 0.01}$ & $0.320_{\pm 0.01}$
& $0.496_{\pm 0.00}$ & $0.158_{\pm 0.01}$ & $0.499_{\pm 0.00}$ & $0.499_{\pm 0.00}$ & $0.496_{\pm 0.00}$ \\
\midrule

\multicolumn{11}{c}{\textit{Chest X Pneumothorax}} \\
ResNet-18 & $0.012_{\pm 0.00}$ & $0.014_{\pm 0.01}$ & $0.027_{\pm 0.00}$ & $0.029_{\pm 0.00}$ & $0.024_{\pm 0.00}$ & $0.495_{\pm 0.00}$ & $0.008_{\pm 0.00}$ & $\bm{0.502_{\pm 0.00}}$ & $\bm{0.502_{\pm 0.00}}$ & $\underline{0.499_{\pm 0.00}}$ \\
ResNet-34 & $0.009_{\pm 0.00}$ & $0.009_{\pm 0.01}$ & $0.027_{\pm 0.00}$ & $\underline{0.030_{\pm 0.01}}$ & $0.022_{\pm 0.00}$ & $0.494_{\pm 0.00}$ & $0.007_{\pm 0.00}$ & $\bm{0.502_{\pm 0.00}}$ & $\bm{0.502_{\pm 0.00}}$ & $\underline{0.499_{\pm 0.00}}$ \\
ResNet-50 & $0.015_{\pm 0.00}$ & $0.017_{\pm 0.00}$ & $0.019_{\pm 0.01}$ & $0.020_{\pm 0.01}$ & $0.020_{\pm 0.00}$ & $0.499_{\pm 0.00}$ & $\underline{0.009_{\pm 0.00}}$ & $\underline{0.501_{\pm 0.00}}$ & $\underline{0.501_{\pm 0.00}}$ & $\bm{0.500_{\pm 0.00}}$ \\
ResNet-50 (IN-1k) & $\underline{0.019_{\pm 0.00}}$ & $0.017_{\pm 0.00}$ & $0.016_{\pm 0.00}$ & $0.018_{\pm 0.00}$ & $0.019_{\pm 0.00}$ & $0.499_{\pm 0.00}$ & $\underline{0.009_{\pm 0.00}}$ & $0.500_{\pm 0.00}$ & $0.500_{\pm 0.00}$ & $\bm{0.500_{\pm 0.00}}$ \\
ResNet-101 & $0.015_{\pm 0.00}$ & $0.013_{\pm 0.00}$ & $0.021_{\pm 0.00}$ & $0.023_{\pm 0.00}$ & $0.020_{\pm 0.00}$ & $0.499_{\pm 0.00}$ & $0.008_{\pm 0.00}$ & $\underline{0.501_{\pm 0.00}}$ & $\underline{0.501_{\pm 0.00}}$ & $\bm{0.500_{\pm 0.00}}$ \\
DenseNet-121 & $0.014_{\pm 0.00}$ & $\bm{0.024_{\pm 0.01}}$ & $\underline{0.030_{\pm 0.00}}$ & $\bm{0.032_{\pm 0.00}}$ & $\bm{0.033_{\pm 0.00}}$ & $0.499_{\pm 0.00}$ & $\bm{0.010_{\pm 0.00}}$ & $\bm{0.502_{\pm 0.00}}$ & $\bm{0.502_{\pm 0.00}}$ & $\bm{0.500_{\pm 0.00}}$ \\
DenseNet-121 (CheXpert) & $0.015_{\pm 0.00}$ & $\underline{0.020_{\pm 0.00}}$ & $0.019_{\pm 0.00}$ & $0.019_{\pm 0.00}$ & $0.014_{\pm 0.00}$ & $\bm{0.502_{\pm 0.00}}$ & $\underline{0.009_{\pm 0.00}}$ & $\bm{0.502_{\pm 0.00}}$ & $\bm{0.502_{\pm 0.00}}$ & $\underline{0.499_{\pm 0.00}}$ \\
DenseNet-169 & $0.014_{\pm 0.00}$ & $\bm{0.024_{\pm 0.01}}$ & $\bm{0.031_{\pm 0.00}}$ & $\bm{0.032_{\pm 0.00}}$ & $\bm{0.033_{\pm 0.00}}$ & $0.496_{\pm 0.00}$ & $\bm{0.010_{\pm 0.00}}$ & $\underline{0.501_{\pm 0.00}}$ & $\bm{0.502_{\pm 0.00}}$ & $\bm{0.500_{\pm 0.00}}$ \\
DenseNet-201 & $0.011_{\pm 0.00}$ & $0.017_{\pm 0.01}$ & $0.028_{\pm 0.00}$ & $\underline{0.030_{\pm 0.00}}$ & $\underline{0.028_{\pm 0.00}}$ & $0.496_{\pm 0.00}$ & $\underline{0.009_{\pm 0.00}}$ & $\bm{0.502_{\pm 0.00}}$ & $\bm{0.502_{\pm 0.00}}$ & $\bm{0.500_{\pm 0.00}}$ \\
ViT-B/16 (224, IN-21k) & $\underline{0.019_{\pm 0.00}}$ & $0.009_{\pm 0.00}$ & $0.024_{\pm 0.01}$ & $0.026_{\pm 0.01}$ & $0.023_{\pm 0.00}$ & $0.500_{\pm 0.00}$ & $0.007_{\pm 0.00}$ & $\underline{0.501_{\pm 0.00}}$ & $\underline{0.501_{\pm 0.00}}$ & $\bm{0.500_{\pm 0.00}}$ \\
ViT-L/16 (224, IN-21k) & $\bm{0.022_{\pm 0.00}}$ & $0.016_{\pm 0.01}$ & $0.021_{\pm 0.00}$ & $0.023_{\pm 0.00}$ & $0.018_{\pm 0.00}$ & $\underline{0.501_{\pm 0.00}}$ & $0.007_{\pm 0.00}$ & $0.500_{\pm 0.00}$ & $0.500_{\pm 0.00}$ & $\bm{0.500_{\pm 0.00}}$ \\
\bottomrule

\end{tabular}
}
\end{table}

\subsection*{Quantitative Evaluation of XAI Methods}
We evaluate explanation quality using the Relevance Rank Accuracy and the proposed Dual-Polarity Precision. Table~\ref{tab:xai_results} reports results averaged over three random seeds.

Across all datasets, RRA and DPP scores remain low for most attribution methods and model variants, despite strong classification performance. On the COVID dataset, the X-ray-pretrained CheXpert model attains the best score in only two of ten comparisons. Larger architectures such as DenseNet-201 and ViT-L/16 often score below smaller counterparts including DenseNet-121 or DenseNet-169 and ViT-B/16. ResNet-101 exceeds smaller ResNet variants only in two of ten comparisons. Taken together, these results do not show a consistent association between increased model size and higher explanation quality on this dataset. In addition, for GradientSHAP, Integrated Gradients, and Saliency, DPP values cluster around 0.5, which corresponds to the baseline and indicates that, although the metric is theoretically sound, it provides limited discriminative signal for these methods in the present setting.

The Oxford-IIIT Pet dataset yields the highest interpretability scores among the evaluated datasets. ResNet-50 pretrained on ImageNet achieves the best result in seven of ten comparisons, with DenseNet-121 and ResNet-50 accounting for the remaining best scores. Although ImageNet pretraining improves performance in this setting, smaller architectures still match or exceed larger models in several comparisons, nevertheless, pretrained ViTs do not show considerable advantage of pretraining over models trained from scratch. For DPP, Grad-CAM produces clearly higher values than the one-sided attribution baseline of 0.0, whereas methods that produce signed attributions remain close to the 0.5 baseline, indicating weak polarity alignment with the ground-truth signal.

On the Chest X Pneumothorax dataset, RRA values are near zero and DPP values are near the baseline scores of 0.5 across models and attribution methods. This pattern indicates poor localization of the relevant pathological regions, even when classification performance is high. One possible explanation is the small spatial extent of the pneumothorax masks. On average, the annotated region covers 1.4\% of the image area, which is far smaller than in the other datasets, where the masks cover 18.1\% for COVID-Qu-Ex and 29.7\% for Oxford-IIIT Pet (Table~\ref{tab:images_masks}). This limited coverage makes the evaluation more sensitive to small localization errors and can depress RRA and DPP even when models capture coarse diagnostic cues. Among the low scores, the DenseNet-121 and DenseNet-169 yield the highest scores among the tested models. The XAI results for the scratch-trained ResNet-50 on the Chest X Pneumothorax dataset should be interpreted cautiously, as this model achieved lower classification performance than the other CNN variants in that setting. Overall, these results suggest that the models may rely on cues outside the annotated regions rather than the intended anatomical evidence. Additional examples are presented in Appendix~\ref{app:pneumo_images_appendix}.

Overall, increasing the number of model parameters does not consistently improve explanation quality under rank (RRA) or mass (DPP) metrics used in this study. Deeper variants such as ResNet-101, DenseNet-201, and ViT-L/16 do not show systematic gains over smaller architectures on the datasets considered in this study. Although three seeds provide only a minimal estimate of inter-seed variability, the standard deviations are consistently small, likely reflecting stable aggregate localisation behaviour across initialisations and averaging over 200 randomly sampled test images per single seed, resulting in 600 evaluated images in total across the three seeds.

\subsection*{Statistical Analysis of Differences in the XAI Results}
We assess whether the observed differences in explanation quality are statistically significant by analysing RRA and DPP scores across all XAI methods and datasets. Experiments are grouped into five comparison settings: ResNet scale (comparing depths 18, 34, 50, and 101), DenseNet scale (comparing depths 121, 169, and 201), Vision Transformer scale (comparing pretrained ViT-B/16 and ViT-L/16) ResNet pretraining (comparing ResNet-50 trained from scratch with an ImageNet-pretrained variant), and DenseNet pretraining (comparing DenseNet-121 trained from scratch with a CheXpert-pretrained variant). Since both Vision Transformers are included only as pretrained models, the ViT-B/16 versus ViT-L/16 comparison should be interpreted as a pretrained-model comparison across scale rather than as a direct analogue of the CNN scale comparisons.

For multi-group comparisons in the ResNet and DenseNet scale settings, we use the Kruskal-Wallis H-test. For binary comparisons in the ViT scale and pretraining settings, we use the Mann-Whitney U-test. All tests are run on explanations computed for 200 randomly sampled test images per seed, resulting in 600 images in total per group. We report effect sizes alongside test results. Cliff’s $\delta$ is used for pairwise tests, and $\eta^2$ is used for multi-group tests, using standard interpretation thresholds. To control the false discovery rate under multiple testing, we adjust all $p$-values with the Benjamini-Hochberg procedure. We estimate statistical power post hoc with bootstrap resampling. For each comparison, we resample the observed group distributions with replacement 1,000 times, re-run the corresponding test, and define power as the fraction of resamples with $p<0.05$. This approach is used because closed-form power calculations are not available for the Mann-Whitney U and Kruskal-Wallis tests. Because power is evaluated at $\alpha=0.05$ rather than the Benjamini-Hochberg adjusted threshold, the resulting estimates are slightly optimistic and interpreted as indicative only. Full results are reported in Table~\ref{tab:stats_all} in Appendix~\ref{app:stat}. Table~\ref{tab:stat_sig_scale} and Table~\ref{tab:stat_sig_pretrain} summarise the full results presented in Table~\ref{tab:stats_all} by reporting only statistically significant comparisons and the corresponding model ordering.

\begin{table}[t!]
\centering
\small
\caption{Statistical comparison of XAI methods across model scales, ranked 1\textsuperscript{st}--4\textsuperscript{th} by Relevance Rank Accuracy (RRA) and Dual-Polarity Precision (DPP) localization scores. $p$-values are Benjamini-Hochberg adjusted. Effect sizes are $\eta^2$ (Kruskal-Wallis) or $\delta$ (pairwise Mann-Whitney U). Lighter cells denote smaller models. Shared superscripts denote tied ranks.}
\vspace{-1em}
\begin{tabular}{lllcccccc}
\toprule
\multirow{2}{*}{\shortstack{Scale \\ Comparison}} & \multirow{2}{*}{Metric} & \multirow{2}{*}{XAI Method} & \multirow{2}{*}{\shortstack{$p$-value \\ (BH-adj.)}} & \multirow{2}{*}{\shortstack{Effect Size \\ ($\eta^2$ or $\delta$)}} & \multicolumn{4}{c}{Rank} \\
\cmidrule{6-9}
& & & & & 1\textsuperscript{st} & 2\textsuperscript{nd} & 3\textsuperscript{rd} & 4\textsuperscript{th} \\
\midrule
\multicolumn{9}{c}{\textit{COVID-QU-Ex}} \\

\multirow{8}{*}{ResNets} & \multirow{3}{*}{RRA} &
Integrated Gradients & 0.04 & 0.00 &
\cellcolor{resnet-34}RN-34 &
\cellcolor{resnet-18}RN-18 &
\cellcolor{resnet-101}RN-101 &
\cellcolor{resnet-50}RN-50 \\
& & Grad-CAM & $<0.01$ & 0.01 &
\cellcolor{resnet-18}RN-18 &
\cellcolor{resnet-101}RN-101 &
\cellcolor{resnet-50}RN-50 &
\cellcolor{resnet-34}RN-34 \\
& & Feature Permutation & $<0.01$ & 0.03 &
\cellcolor{resnet-34}RN-34 &
\cellcolor{resnet-18}RN-18 &
\cellcolor{resnet-101}RN-101 &
\cellcolor{resnet-50}RN-50 \\
\cmidrule{2-9}
& \multirow{5}{*}{DPP} &
Saliency & $<0.01$ & 0.02 &
\cellcolor{resnet-18}RN-18\textsuperscript{a} &
\cellcolor{resnet-34}RN-34\textsuperscript{a} &
\cellcolor{resnet-50}RN-50\textsuperscript{b} &
\cellcolor{resnet-101}RN-101\textsuperscript{b} \\
& & Integrated Gradients & $<0.01$ & 0.02 &
\cellcolor{resnet-101}RN-101 &
\cellcolor{resnet-18}RN-18 &
\cellcolor{resnet-34}RN-34\textsuperscript{a} &
\cellcolor{resnet-50}RN-50\textsuperscript{a} \\
& & GradientSHAP & $<0.01$ & 0.01 &
\cellcolor{resnet-101}RN-101 &
\cellcolor{resnet-18}RN-18 &
\cellcolor{resnet-50}RN-50 &
\cellcolor{resnet-34}RN-34 \\
& & Grad-CAM & 0.03 & 0.00 &
\cellcolor{resnet-18}RN-18 &
\cellcolor{resnet-101}RN-101 &
\cellcolor{resnet-34}RN-34\textsuperscript{a} &
\cellcolor{resnet-50}RN-50\textsuperscript{a} \\
& & Feature Permutation & $<0.01$ & 0.02 &
\cellcolor{resnet-34}RN-34 &
\cellcolor{resnet-18}RN-18 & 
\cellcolor{resnet-101}RN-101 &
\cellcolor{resnet-50}RN-50 \\
\midrule

\multirow{9}{*}{DenseNets} & \multirow{5}{*}{RRA} &
Saliency & $<0.01$ & 0.02 &
\cellcolor{resnet-34}DN-169 &
\cellcolor{resnet-18}DN-121 &
\cellcolor{resnet-50}DN-201 &
-- \\
& & Integrated Gradients & 0.04 & 0.00 &
\cellcolor{resnet-34}DN-169 &
\cellcolor{resnet-18}DN-121 &
\cellcolor{resnet-50}DN-201 &
-- \\
& & GradientSHAP & 0.03 & 0.00 &
\cellcolor{resnet-34}DN-169 &
\cellcolor{resnet-18}DN-121 &
\cellcolor{resnet-50}DN-201 &
-- \\
& & Grad-CAM & $<0.01$ & 0.05 &
\cellcolor{resnet-50}DN-201 &
\cellcolor{resnet-34}DN-169 &
\cellcolor{resnet-18}DN-121 &
-- \\
& & Feature Permutation & $<0.01$ & 0.01 &
\cellcolor{resnet-34}DN-169 &
\cellcolor{resnet-50}DN-201 &
\cellcolor{resnet-18}DN-121 &
-- \\
\cmidrule{2-9}
& \multirow{4}{*}{DPP} &
Saliency & $<0.01$ & 0.01 &
\cellcolor{resnet-18}DN-121\textsuperscript{a} &
\cellcolor{resnet-50}DN-201\textsuperscript{a} &
\cellcolor{resnet-34}DN-169 &
-- \\
& & Integrated Gradients & $<0.01$ & 0.01 &
\cellcolor{resnet-50}DN-201 &
\cellcolor{resnet-34}DN-169 &
\cellcolor{resnet-18}DN-121 &
-- \\
& & GradientSHAP & $<0.01$ & 0.01 &
\cellcolor{resnet-34}DN-169\textsuperscript{a} &
\cellcolor{resnet-50}DN-201\textsuperscript{a} &
\cellcolor{resnet-18}DN-121 &
-- \\
& & Grad-CAM & $<0.01$ & 0.03 &
\cellcolor{resnet-50}DN-201 &
\cellcolor{resnet-34}DN-169 &
\cellcolor{resnet-18}DN-121 &
-- \\
\midrule

\multirow{4.5}{*}{ViTs (pretrined)} & RRA &
Saliency & 0.03 & 0.08 &
\cellcolor{resnet-18}ViT-B/16 &
\cellcolor{resnet-34}ViT-L/16 &
-- & -- \\
\cmidrule{2-9}
& \multirow{3}{*}{DPP} &
Saliency & $<0.01$ & -0.10 &
\cellcolor{resnet-34}ViT-L/16 &
\cellcolor{resnet-18}ViT-B/16 &
-- & -- \\
& & Integrated Gradients & $<0.01$ & -0.16 &
\cellcolor{resnet-34}ViT-L/16 &
\cellcolor{resnet-18}ViT-B/16 &
-- & -- \\
& & GradientSHAP & $<0.01$ & -0.15 &
\cellcolor{resnet-34}ViT-L/16 &
\cellcolor{resnet-18}ViT-B/16 &
-- & -- \\

\bottomrule
\multicolumn{9}{c}{\textit{Oxford-IIIT Pet}} \\
\multirow{3}{*}{ResNets} & RRA &
Grad-CAM & $<0.01$ & 0.01 &
\cellcolor{resnet-101}RN-101 &
\cellcolor{resnet-50}RN-50 &
\cellcolor{resnet-18}RN-18 & 
\cellcolor{resnet-34}RN-34 \\
\cmidrule{2-9}
& \multirow{2}{*}{DPP} &
Grad-CAM & 0.01 & 0.01 &
\cellcolor{resnet-101}RN-101 &
\cellcolor{resnet-18}RN-18\textsuperscript{a} &
\cellcolor{resnet-50}RN-50\textsuperscript{a} &
\cellcolor{resnet-34}RN-34 \\
& & Feature Permutation & 0.01 & 0.00 &
\cellcolor{resnet-50}RN-50 &
\cellcolor{resnet-101}RN-101 &
\cellcolor{resnet-18}RN-18 &
\cellcolor{resnet-34}RN-34 \\
\midrule

\multirow{2}{*}{DenseNets} & RRA &
Feature Permutation & 0.04 & 0.00 &
\cellcolor{resnet-34}DN-169 &
\cellcolor{resnet-50}DN-201 &
\cellcolor{resnet-18}DN-121 &
-- \\
\cmidrule{2-9}
& DPP &
Feature Permutation & $<0.01$ & 0.01 &
\cellcolor{resnet-34}DN-169 &
\cellcolor{resnet-50}DN-201 &
\cellcolor{resnet-18}DN-121 &
-- \\
\midrule

\multirow{3}{*}{ViTs (pretrained)} & \multirow{3}{*}{RRA} &
Saliency & 0.04 & 0.08 &
\cellcolor{resnet-18}ViT-B/16 &
\cellcolor{resnet-34}ViT-L/16 &
-- & -- \\
& & Integrated Gradients & 0.01 & 0.10 &
\cellcolor{resnet-18}ViT-B/16 &
\cellcolor{resnet-34}ViT-L/16 &
-- & -- \\
& & GradientSHAP & <0.01 & 0.10 &
\cellcolor{resnet-18}ViT-B/16 &
\cellcolor{resnet-34}ViT-L/16 &
-- & -- \\ 

\bottomrule
\multicolumn{8}{c}{\textit{Chest X Pneumothorax}} \\
\multirow{8}{*}{ResNets} & \multirow{4}{*}{RRA} &
Saliency & $<0.01$ & 0.01 &
\cellcolor{resnet-18}RN-18 &
\cellcolor{resnet-34}RN-34 &
\cellcolor{resnet-50}RN-50\textsuperscript{a} &
\cellcolor{resnet-101}RN-101\textsuperscript{a} \\
& & Integrated Gradients & $<0.01$ & 0.02 &
\cellcolor{resnet-34}RN-34 &
\cellcolor{resnet-18}RN-18 &
\cellcolor{resnet-101}RN-101 &
\cellcolor{resnet-50}RN-50 \\
& & GradientSHAP & $<0.01$ & 0.01 &
\cellcolor{resnet-18}RN-18\textsuperscript{a} &
\cellcolor{resnet-34}RN-34\textsuperscript{a} &
\cellcolor{resnet-101}RN-101 &
\cellcolor{resnet-50}RN-50 \\
& & Feature Permutation & 0.01 & 0.00 &
\cellcolor{resnet-101}RN-101\textsuperscript{a} &
\cellcolor{resnet-50}RN-50\textsuperscript{a} &
\cellcolor{resnet-18}RN-18 &
\cellcolor{resnet-34}RN-34 \\
\cmidrule{2-9}
& \multirow{4}{*}{DPP} &
Saliency & $<0.01$ & 0.00 &
\cellcolor{resnet-50}RN-50\textsuperscript{a} &
\cellcolor{resnet-101}RN-101\textsuperscript{a} &
\cellcolor{resnet-18}RN-18\textsuperscript{b} &
\cellcolor{resnet-34}RN-34\textsuperscript{b} \\
& & Integrated Gradients & $<0.01$ & 0.02 &
\cellcolor{resnet-18}RN-18\textsuperscript{a} &
\cellcolor{resnet-34}RN-34\textsuperscript{a} &
\cellcolor{resnet-50}RN-50\textsuperscript{b} &
\cellcolor{resnet-101}RN-101\textsuperscript{b} \\
& & GradientSHAP & $<0.01$ & 0.02 &
\cellcolor{resnet-18}RN-18\textsuperscript{a} &
\cellcolor{resnet-34}RN-34\textsuperscript{a} &
\cellcolor{resnet-50}RN-50\textsuperscript{b} &
\cellcolor{resnet-101}RN-101\textsuperscript{b} \\
& & Feature Permutation & $<0.01$ & 0.02 &
\cellcolor{resnet-50}RN-50\textsuperscript{a} &
\cellcolor{resnet-101}RN-101\textsuperscript{a} &
\cellcolor{resnet-18}RN-18 &
\cellcolor{resnet-34}RN-34 \\
\midrule

\multirow{2}{*}{DenseNets} & \multirow{2}{*}{DPP} &
Integrated Gradients & 0.04 & 0.00 &
\cellcolor{resnet-18}DN-121\textsuperscript{a} &
\cellcolor{resnet-34}DN-169\textsuperscript{a} &
\cellcolor{resnet-50}DN-201 &
-- \\
& & Feature Permutation & $<0.01$ & 0.02 &
\cellcolor{resnet-18}DN-121\textsuperscript{a} &
\cellcolor{resnet-34}DN-169\textsuperscript{a} &
\cellcolor{resnet-50}DN-201 &
-- \\
\midrule

\multirow{3.5}{*}{ViTs (pretrained)} & \multirow{2}{*}{RRA} &
Saliency & 0.02 & 0.09 &
\cellcolor{resnet-18}ViT-B/16 &
\cellcolor{resnet-34}ViT-L/16 &
-- & -- \\
& & Grad-CAM & $<0.01$ & -0.17 &
\cellcolor{resnet-34}ViT-L/16 &
\cellcolor{resnet-18}ViT-B/16 &
-- & --\\

\cmidrule{2-9}
& DPP &
Integrated Gradients & 0.01 & 0.10 &
\cellcolor{resnet-18}ViT-B/16 &
\cellcolor{resnet-34}ViT-L/16 &
-- & -- \\

\bottomrule
\end{tabular}
\label{tab:stat_sig_scale}
\end{table}

\subsubsection*{Analysis of Model Scale Effect}
We next describe results that differ significantly between model scales, followed by results with no detected differences. Table~\ref{tab:stat_sig_scale} summarises the XAI methods for which scale-dependent differences are observed. Across 90 tests assessing the effect of model scale on explanation quality, 42 tests (47\%) yield statistically significant differences after correction. In 11 of 90 comparisons (12\%), the largest model within a family achieves the highest score. In the remaining 31 significant comparisons (34\%), smaller models match or exceed the performance of the largest model. For 48 of 90 tests (53\%), we do not reject the null hypothesis of equal explanation quality across model scales. Nevertheless, the associated effect sizes are typically negligible to small, indicating that even when differences are statistically significant, they are often limited in practical importance.

\begin{table}[h!]
\centering
\small
\caption{Statistical comparison of XAI methods between scratch-trained and pretrained variants, ranked 1\textsuperscript{st}--2\textsuperscript{nd} by Relevance Rank Accuracy (RRA) and Dual-Polarity Precision (DPP) localization scores. $p$-values are Benjamini-Hochberg adjusted. Effect sizes are $\delta$ (pairwise Mann-Whitney U). Lighter cells denote scratch-trained models, darker denote pretrained versions. Shared superscripts denote tied ranks. Note: \textsuperscript{$\dagger$} indicates model that did not converge within the training budget of 100 epochs across all seeds.}
\begin{tabular}{lllcccc}
\toprule
\multirow{2}{*}{\shortstack{Pretraining \\ Comparison}} & \multirow{2}{*}{Metric} & \multirow{2}{*}{XAI Method} & \multirow{2}{*}{\shortstack{$p$-value \\ (BH-adj.)}} & \multirow{2}{*}{\shortstack{Effect Size \\ ($\delta$)}} & \multicolumn{2}{c}{Rank} \\
\cmidrule{6-7}
& & & & & 1\textsuperscript{st} & 2\textsuperscript{nd} \\
\midrule
\multicolumn{7}{c}{\textit{COVID-QU-Ex}} \\

\multirow{6}{*}{ResNets} & \multirow{3}{*}{RRA} &
Saliency & $<0.01$ & 0.25 &
\cellcolor{resnet-18}RN-50 &
\cellcolor{resnet-34}RN-50 IN-1k \\
& & GradientSHAP & 0.02 & 0.09 &
\cellcolor{resnet-18}RN-50 &
\cellcolor{resnet-34}RN-50 IN-1k \\
& & Grad-CAM & $<0.01$ & -0.14 &
\cellcolor{resnet-34}RN-50 IN-1k &
\cellcolor{resnet-18}RN-50 \\
\cmidrule{2-7}
& \multirow{3}{*}{DPP} &
Saliency & $<0.01$ & -0.16 &
\cellcolor{resnet-34}RN-50 IN-1k &
\cellcolor{resnet-18}RN-50 \\
& & Integrated Gradients & $<0.01$ & -0.12 &
\cellcolor{resnet-34}RN-50 IN-1k &
\cellcolor{resnet-18}RN-50 \\
& & Grad-CAM & 0.03 & -0.08 &
\cellcolor{resnet-34}RN-50 IN-1k &
\cellcolor{resnet-18}RN-50 \\
\midrule

\multirow{8}{*}{DenseNets} & \multirow{4}{*}{RRA} &
Integrated Gradients & 0.03 & -0.08 &
\cellcolor{resnet-34}DN-121 CheXpert\textsuperscript{$\dagger$} &
\cellcolor{resnet-18}DN-121 \\
& & GradientSHAP & 0.03 & -0.08 &
\cellcolor{resnet-34}DN-121 CheXpert\textsuperscript{$\dagger$} &
\cellcolor{resnet-18}DN-121 \\
& & Grad-CAM & $<0.01$ & -0.34 &
\cellcolor{resnet-34}DN-121 CheXpert\textsuperscript{$\dagger$} &
\cellcolor{resnet-18}DN-121 \\
& & Feature Permutation & $<0.01$ & 0.22 &
\cellcolor{resnet-18}DN-121 & 
\cellcolor{resnet-34}DN-121 CheXpert\textsuperscript{$\dagger$} \\
\cmidrule{2-7}
& \multirow{4}{*}{DPP} &
Integrated Gradients & 0.01 & -0.10 &
\cellcolor{resnet-34}DN-121 CheXpert\textsuperscript{$\dagger$} &
\cellcolor{resnet-18}DN-121 \\
& & GradientSHAP & 0.01 & -0.09 &
\cellcolor{resnet-34}DN-121 CheXpert\textsuperscript{$\dagger$} &
\cellcolor{resnet-18}DN-121 \\
& & Grad-CAM & $<0.01$ & -0.28 &
\cellcolor{resnet-34}DN-121 CheXpert\textsuperscript{$\dagger$} &
\cellcolor{resnet-18}DN-121 \\
& & Feature Permutation & $<0.01$ & 0.54 &
\cellcolor{resnet-18}DN-121 &
\cellcolor{resnet-34}DN-121 CheXpert\textsuperscript{$\dagger$} \\
\bottomrule

\multicolumn{7}{c}{\textit{Oxford-IIIT Pet}} \\
\multirow{4}{*}{ResNets} & \multirow{3}{*}{RRA} &
Saliency & $<0.01$ & -0.13 &
\cellcolor{resnet-34}RN-50 IN-1k &
\cellcolor{resnet-18}RN-50 \\ 
& & Grad-CAM & $<0.01$ & -0.20 &
\cellcolor{resnet-34}RN-50 IN-1k &
\cellcolor{resnet-18}RN-50 \\
& & Feature Permutation & $<0.01$ & -0.13 &
\cellcolor{resnet-34}RN-50 IN-1k &
\cellcolor{resnet-18}RN-50 \\
\cmidrule{2-7}
& DPP &
Grad-CAM & $<0.01$ & -0.21 &
\cellcolor{resnet-34}RN-50 IN-1k &
\cellcolor{resnet-18}RN-50 \\
\midrule

\multirow{6}{*}{DenseNets} & \multirow{3}{*}{RRA} &
Saliency & $<0.01$ & 0.11 &
\cellcolor{resnet-18}DN-121 &
\cellcolor{resnet-34}DN-121 CheXpert\textsuperscript{$\dagger$} \\
& & Integrated Gradients & $<0.01$ & 0.11 &
\cellcolor{resnet-18}DN-121 &
\cellcolor{resnet-34}DN-121 CheXpert\textsuperscript{$\dagger$} \\
& & GradientSHAP & 0.01 & 0.10 &
\cellcolor{resnet-18}DN-121 &
\cellcolor{resnet-34}DN-121 CheXpert\textsuperscript{$\dagger$} \\
\cmidrule{2-7}
& \multirow{3}{*}{DPP} &
Integrated Gradients & $<0.01$ & 0.13 &
\cellcolor{resnet-18}DN-121 &
\cellcolor{resnet-34}DN-121 CheXpert\textsuperscript{$\dagger$} \\
& & GradientSHAP & $<0.01$ & 0.13 &
\cellcolor{resnet-18}DN-121 &
\cellcolor{resnet-34}DN-121 CheXpert\textsuperscript{$\dagger$} \\
& & Feature Permutation & $<0.01$ & -0.20 &
\cellcolor{resnet-34}DN-121 CheXpert\textsuperscript{$\dagger$} &
\cellcolor{resnet-18}DN-121 \\
\bottomrule

\multicolumn{7}{c}{\textit{Chest X Pneumothorax}} \\
\multirow{2}{*}{ResNets} & RRA &
Integrated Gradients & 0.01 & -0.09 &
\cellcolor{resnet-18}RN-50 & 
\cellcolor{resnet-34}RN-50 IN-1k \\ 
\cmidrule{2-7}
& DPP &
Saliency & 0.01 & -0.10 &
\cellcolor{resnet-18}RN-50\textsuperscript{a} &
\cellcolor{resnet-34}RN-50 IN-1k\textsuperscript{a} \\
\midrule

\multirow{4}{*}{DenseNets} & \multirow{3}{*}{DPP} &
Saliency & $<0.01$ & 0.31 &
\cellcolor{resnet-18}DN-121 &
\cellcolor{resnet-34}DN-121 CheXpert \\
& & Integrated Gradients & $<0.01$ & 0.15 &
\cellcolor{resnet-18}DN-121 &
\cellcolor{resnet-34}DN-121 CheXpert \\
& & GradientSHAP & $<0.01$ & 0.16 &
\cellcolor{resnet-18}DN-121 &
\cellcolor{resnet-34}DN-121 CheXpert \\ 
\cmidrule{2-7}
& DPP &
Saliency & 0.01 & 0.11 &
\cellcolor{resnet-18}DN-121 &
\cellcolor{resnet-34}DN-121 CheXpert \\

\bottomrule
\end{tabular}
\label{tab:stat_sig_pretrain}
\end{table}

\begin{table}[!htbp]
\centering
\caption{Sanity check analysis on the Oxford-IIIT Pet dataset. We evaluate Layer-wise Relevance Randomization via Mean Degradation (where higher $\uparrow$ indicates dependence on trained weights) and Input Randomization via Mean Sensitivity (where higher $\uparrow$ indicates responsiveness to input changes). Cliff's $\delta$ effect sizes are provided to interpret the magnitude of changes in explanation maps under input randomization.}
\label{tab:sanity_checks_summary}

\begin{subtable}[t]{\textwidth}
\centering
\caption{Scale sanity checks across selected ResNet, DenseNet, and pretrained Vision Transformer models}
\label{tab:sanity_checks_summary_sacle}
\resizebox{\textwidth}{!}{
\begin{tabular}{llccr}
\toprule
Model & Method &
\makecell{Layer-wise Relevance\\ Randomisation (Degradation) $\uparrow$} &
\makecell{Input Randomisation\\ (Sensitivity) $\uparrow$} &
\makecell{Input Randomisation\\ Cliff's $\delta$ (Interp.)} \\
\midrule
\multirow{5}{*}{ResNet-18}
& Saliency             & $0.788_{\pm 0.05}$ & $0.007_{\pm 0.00}$ & $0.12$ (Negligible) \\
& Integrated Gradients & $0.688_{\pm 0.01}$ & $0.003_{\pm 0.00}$ & $-0.41$ (Medium) \hspace{0.565em} \\
& GradientSHAP         & $0.773_{\pm 0.01}$ & $0.004_{\pm 0.00}$ & $-0.20$ (Small) \hspace{1.59em} \\
& Grad-CAM              & $0.619_{\pm 0.03}$ & $0.082_{\pm 0.04}$ & $-0.01$ (Negligible) \\
& Feature Permutation  & $0.762_{\pm 0.06}$ & $0.017_{\pm 0.01}$ & $-0.20$ (Small) \hspace{1.59em} \\

\midrule
\multirow{5}{*}{ResNet-101}
& Saliency             & $0.890_{\pm 0.01}$ & $0.010_{\pm 0.00}$ & $0.15$ (Negligible) \\
& Integrated Gradients & $0.803_{\pm 0.04}$ & $0.003_{\pm 0.00}$ & $-0.28$ (Small) \hspace{1.59em} \\
& GradientSHAP         & $0.888_{\pm 0.03}$ & $0.005_{\pm 0.00}$ & $-0.47$ (Medium) \hspace{0.565em} \\
& Grad-CAM              & $0.575_{\pm 0.08}$ & $0.101_{\pm 0.08}$ & $-0.04$ (Negligible) \\
& Feature Permutation  & $0.934_{\pm 0.02}$ & $0.012_{\pm 0.00}$ & $-0.07$ (Negligible) \\

\midrule
\multirow{5}{*}{DenseNet-121}
& Saliency             & $0.729_{\pm 0.01}$ & $0.012_{\pm 0.00}$ & $0.09$ (Negligible) \\
& Integrated Gradients & $0.668_{\pm 0.03}$ & $0.004_{\pm 0.00}$ & $-0.23$ (Small) \hspace{1.59em} \\
& GradientSHAP         & $0.766_{\pm 0.02}$ & $0.005_{\pm 0.00}$ & $0.20$ (Small) \hspace{1.59em} \\
& Grad-CAM             & $0.534_{\pm 0.04}$ & $0.048_{\pm 0.02}$ & $0.17$ (Small) \hspace{1.59em} \\
& Feature Permutation  & $0.560_{\pm 0.06}$ & $0.078_{\pm 0.02}$ & $-0.44$ (Medium) \hspace{0.565em} \\

\midrule
\multirow{5}{*}{DenseNet-201}
& Saliency             & $0.740_{\pm 0.04}$ & $0.015_{\pm 0.00}$ & $0.33$ (Medium) \hspace{0.565em} \\
& Integrated Gradients & $0.722_{\pm 0.08}$ & $0.004_{\pm 0.00}$ & $0.71$ (Large) \hspace{1.64em} \\
& GradientSHAP         & $0.814_{\pm 0.05}$ & $0.005_{\pm 0.00}$ & $0.07$ (Negligible) \\
& Grad-CAM             & $0.570_{\pm 0.05}$ & $0.069_{\pm 0.04}$ & $0.15$ (Negligible) \\
& Feature Permutation  & $0.714_{\pm 0.09}$ & $0.067_{\pm 0.02}$ & $0.04$ (Negligible) \\

\midrule
\multirow{5}{*}{ViT-B/16 (224, IN-21k)}
& Saliency             & $0.832_{\pm 0.03}$ & $0.030_{\pm 0.01}$ & $-0.47$ (Medium) \hspace{0.565em} \\
& Integrated Gradients & $0.800_{\pm 0.01}$ & $0.008_{\pm 0.00}$ & $0.15$ (Negligible) \\
& GradientSHAP         & $0.930_{\pm 0.01}$ & $0.013_{\pm 0.01}$ & $-0.15$ (Negligible) \\
& Grad-CAM             & $0.693_{\pm 0.05}$ & $0.000_{\pm 0.00}$ & $-0.09$ (Negligible) \\
& Feature Permutation  & $0.929_{\pm 0.01}$ & $0.010_{\pm 0.00}$ & $0.25$ (Small) \hspace{1.59em} \\

\midrule
\multirow{5}{*}{ViT-L/16 (224, IN-21k)}
& Saliency             & $0.785_{\pm 0.05}$ & $0.020_{\pm 0.01}$ & $-0.20$ (Small) \hspace{1.59em} \\
& Integrated Gradients & $0.801_{\pm 0.03}$ & $0.008_{\pm 0.00}$ & $-0.01$ (Negligible) \\
& GradientSHAP         & $0.922_{\pm 0.03}$ & $0.013_{\pm 0.01}$ & $0.12$ (Negligible) \\
& Grad-CAM             & $0.571_{\pm 0.09}$ & $0.000_{\pm 0.00}$ & $-0.15$ (Negligible) \\
& Feature Permutation  & $0.880_{\pm 0.01}$ & $0.006_{\pm 0.00}$ & $-0.15$ (Negligible) \\

\bottomrule
\end{tabular}
}
\end{subtable}


\centering
\begin{subtable}[h]{\textwidth}
\centering
\vspace{1em}
\caption{Pretraining sanity check for ResNet-50 trained from scratch and ImageNet-pretrained ResNet-50}
\label{tab:sanity_checks_summary_pretrain}
\resizebox{\textwidth}{!}{
\begin{tabular}{llccr}
\toprule
Model & Method &
\makecell{Layer-wise Relevance\\ Randomisation (Degradation) $\uparrow$} &
\makecell{Input Randomisation\\ (Sensitivity) $\uparrow$} &
\makecell{Input Randomisation\\ Cliff's $\delta$ (Interp.)} \\
\midrule
\multirow{5}{*}{ResNet-50}
& Saliency             & $0.848_{\pm 0.02}$ & $0.003_{\pm 0.00}$ & $-0.20$ (Small) \hspace{1.59em} \\
& Integrated Gradients & $0.713_{\pm 0.04}$ & $0.001_{\pm 0.00}$ & $-0.15$ (Negligible) \\
& GradientSHAP         & $0.801_{\pm 0.03}$ & $0.002_{\pm 0.00}$ & $-0.20$ (Small) \hspace{1.59em} \\
& Grad-CAM              & $0.593_{\pm 0.03}$ & $0.026_{\pm 0.01}$ & $-0.17$ (Small) \hspace{1.59em} \\
& Feature Permutation  & $0.857_{\pm 0.05}$ & $0.006_{\pm 0.00}$ & $-0.31$ (Small) \hspace{1.59em} \\
\midrule
\multirow{5}{*}{ResNet-50 (IN-1k)}
& Saliency             & $0.998_{\pm 0.00}$ & $0.082_{\pm 0.00}$ & $-0.04$ (Negligible) \\
& Integrated Gradients & $0.994_{\pm 0.00}$ & $0.019_{\pm 0.00}$ & $-0.25$ (Small) \hspace{1.59em} \\
& GradientSHAP         & $0.996_{\pm 0.00}$ & $0.033_{\pm 0.00}$ & $-0.01$ (Negligible) \\
& Grad-CAM              & $0.904_{\pm 0.01}$ & $0.030_{\pm 0.01}$ & $0.12$ (Negligible) \\
& Feature Permutation  & $0.986_{\pm 0.00}$ & $0.028_{\pm 0.00}$ & $-0.04$ (Negligible) \\
\bottomrule
\end{tabular}
}
\end{subtable}
\end{table}

\subsubsection*{Analysis of Pretraining Effects}
Table~\ref{tab:stat_sig_pretrain} summarises the XAI methods for which differences are observed between models trained from scratch and pretrained models. Across 60 tests assessing the effect of pretraining on explanation quality, 30 tests (50\%) yield statistically significant differences after correction. In 15 of 60 comparisons (25\%), pretrained models achieve higher scores than their scratch-trained counterparts. In the other 15 significant comparisons (25\%), scratch-trained models match or exceed the pretrained models. For the remaining 30 tests (50\%), we do not reject the null hypothesis of equal explanation quality between the training variants. The observed effect sizes are also mostly negligible to small. For COVID-QU-Ex and Oxford-IIIT Pet, however, DenseNet pretraining results involving the CheXpert-pretrained DenseNet-121 are marked with \textsuperscript{$\dagger$} and should be interpreted for completeness rather than as primary evidence for the effect of pretraining, since this model did not satisfy the convergence criterion on those datasets.

\subsection*{Fidelity Tests for Explanations}
We evaluate explanation fidelity using two sanity-check procedures applied to trained models. The first test measures dependence on learned parameters through layer-wise randomisation. We reinitialise the weights in ten randomly selected layers and compare explanations from the perturbed model with those from the original model. We define degradation as one minus the absolute correlation between the two explanation maps. Larger values indicate that weight randomisation changes the explanation, which indicates dependence on learned parameters rather than network structure alone.

The second test measures sensitivity to input structure through input randomisation. We shuffle pixels in the test images and compute explanations for both the original and shuffled inputs. We quantify change using the $\ell_2$ distance between the corresponding attribution maps and report Cliff’s $\delta$ as an effect size for differences between the distance distributions. Under this test, explanations are expected to change when the input content is disrupted by pixel shuffling.

Table~\ref{tab:sanity_checks_summary_sacle} reports results for selected ResNet and DenseNet models together with pretrained ViT-B/16 and ViT-L/16, and Table~\ref{tab:sanity_checks_summary_pretrain} reports results for ResNet-50 trained from scratch and a ImageNet-pretrained variant. All metrics are computed on a batch of 32 randomly drawn test images per single seed from the Oxford-IIIT Pet dataset. We use this dataset because it yields the highest interpretability scores in our experiments and therefore provides favourable conditions for explanation quality. If a method fails these tests under these conditions, it is likely to fail on datasets where localisation performance is lower. For Vision Transformers, the sanity-check analysis is limited to pretrained models and should therefore be interpreted as a pretrained-model comparison across scale.

\subsubsection*{Sanity Checks for Models Differing in Scale}
Layer-wise randomisation yields moderate to high degradation across methods, indicating dependence on learned parameters. For the selected ResNet models, degradation is higher for ResNet-101 than for ResNet-18 for all methods except Grad-CAM. A similar pattern is observed for the DenseNet models, where DenseNet-201 shows higher degradation than DenseNet-121 for all tested XAI methods. For the pretrained Vision Transformers, degradation is also generally high, with ViT-B/16 exceeding ViT-L/16 for Saliency, GradientSHAP, Grad-CAM, and Feature Permutation, whereas Integrated Gradients remains nearly unchanged across the two models.

Under pixel permutation, most gradient-based methods show low sensitivity, with attribution maps changing only slightly when spatial structure is removed. For ResNets, Grad-CAM show larger shifts than the other methods. For the DenseNet models, the largest changes appear for Grad-CAM and Feature Permutation. For the pretrained Vision Transformers, input sensitivity remains low for most methods, with Grad-CAM near zero for both ViT-B/16 and ViT-L/16, while Saliency and GradientSHAP show slightly larger changes.

Cliff’s $\delta$ is usually negligible to small across methods and architectures, with only six of the 30 comparisons reaching the medium or large range. For the ResNet models, medium effects occur for Integrated Gradients on ResNet-18 and for GradientSHAP on ResNet-101. For the DenseNet models, Feature Permutation on DenseNet-121 and Saliency on DenseNet-201 show medium effects, while Integrated Gradients on DenseNet-201 shows a large effect. For the pretrained Vision Transformers, most effects remain negligible to small, with a medium effect observed only for Saliency on ViT-B/16. Overall, differences between architectures follow similar patterns across methods.

\begin{figure*}[t!]
\centering
\begin{subfigure}[t]{\textwidth}
  \centering
  \subcaption{\textbf{Layer-wise weight randomisation.} Explanations for an image using the original model (top) and after layer randomisation (bottom). Coefficient $r$ shows the correlation between the attribution maps generated by the original and perturbed model.}
  \includegraphics[width=\linewidth]{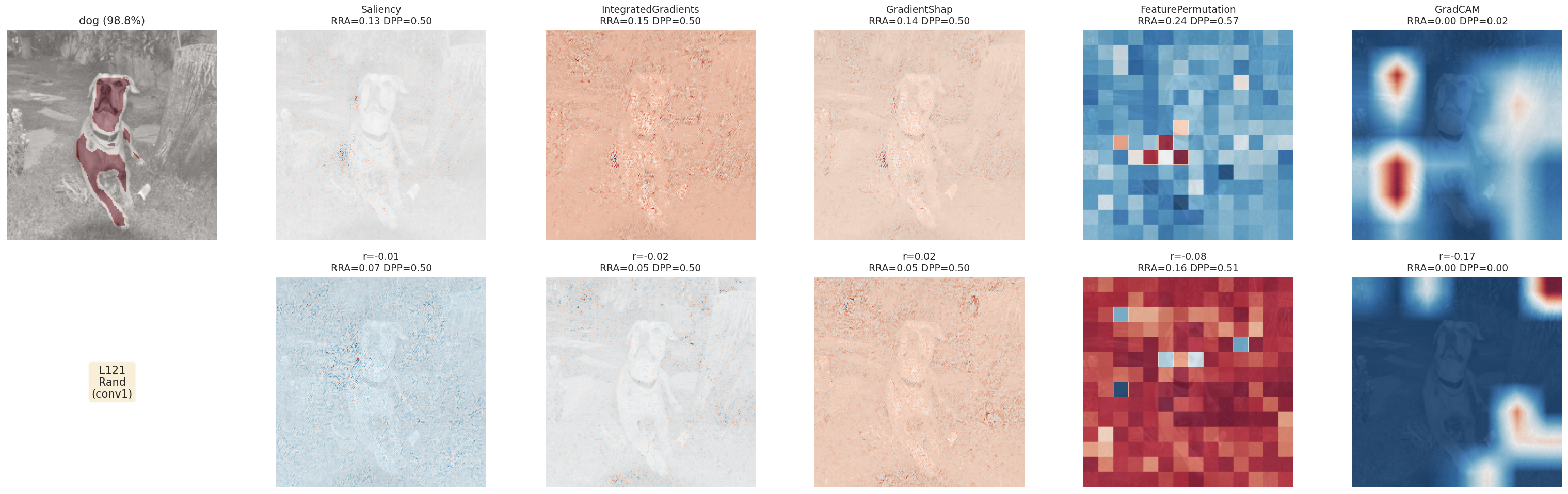}
\label{fig:sanity_pretrained_layerwise}
\end{subfigure}

\begin{subfigure}[t]{\textwidth}
  \centering
  \subcaption{\textbf{Input (pixel) randomisation.} Explanations for two image samples using the original input (top) and a pixel-permuted input (bottom). The $\ell_2$ values are included, presenting the distance between the attribution maps from the original and perturbed input image.}
  \includegraphics[width=\linewidth]{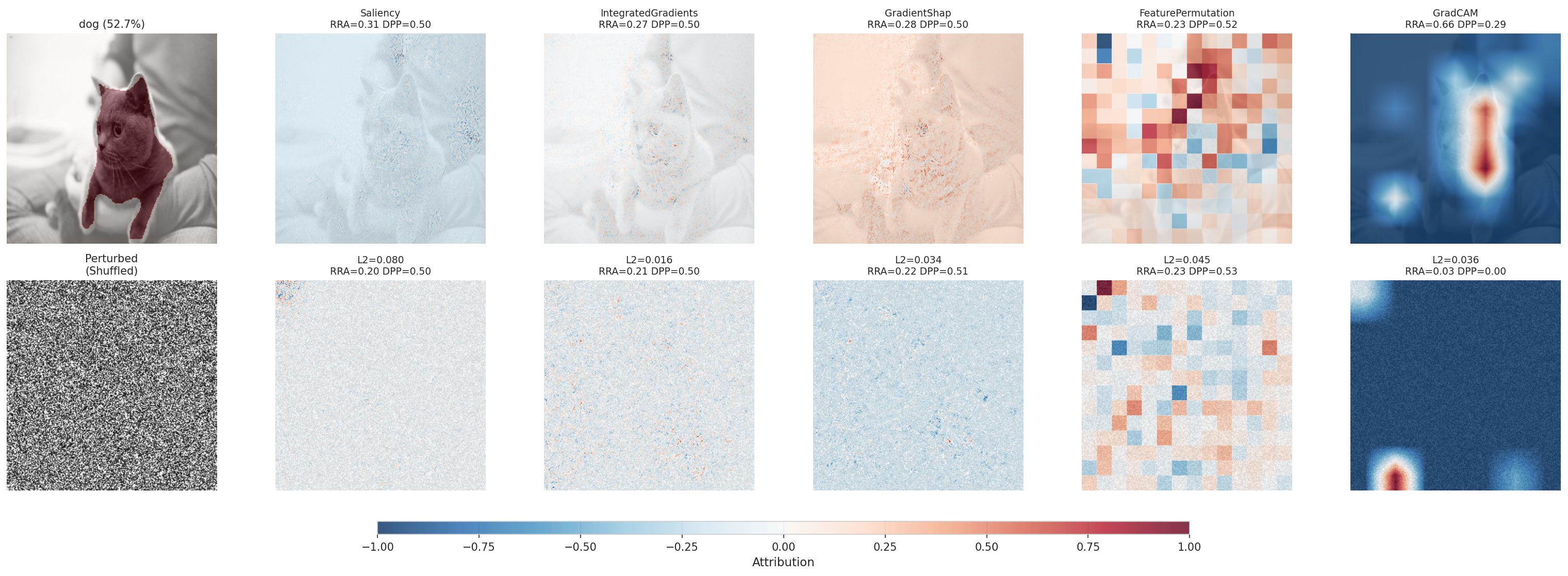}
\label{fig:sanity_pretrained_input}
\end{subfigure}
\caption{Sanity checks for an ImageNet-pretrained ResNet-50 across five XAI methods. Columns show attribution maps from Saliency, Integrated Gradients, GradientSHAP, Feature Permutation, and Grad-CAM. For each method, the scores above the maps report DPP and RRA, and the colour bar indicates normalised attribution magnitude. Panels (a) and (b) define the weight and input randomisation settings and report the corresponding similarity measures.}
\label{fig:sanity_checks}
\end{figure*}

\subsubsection*{Sanity Checks for Trained vs Pretrained Models}
We compare ResNet-50 trained from scratch with an ImageNet-pretrained variant. The pretrained model shows higher degradation under layer-wise randomisation, indicating stronger dependence of the explanations on learned representations and a larger loss of structure after weight reinitialisation.

Sensitivity to pixel permutation increases with pretraining. The largest increase occurs for Saliency, from 0.003 to 0.082. Cliff’s $\delta$ shifts from mostly small negative effects in the scratch-trained model to negligible negative effects for Saliency and Grad-CAM in the pretrained model, while remaining low overall.

Figure~\ref{fig:sanity_checks} illustrates example outputs under layer and input perturbations. Panel~\ref{fig:sanity_pretrained_layerwise} shows the effect of layer randomisation, Panel~\ref{fig:sanity_pretrained_input} shows the effect of input randomisation. Presented in Panel~\ref{fig:sanity_pretrained_layerwise}, Saliency, Integrated Gradients, and GradientSHAP methods produce visually noisy maps, however, RRA decreases after perturbations, while DPP remains close to 0.5, indicating the assignment of positive and negative attributions similar to a random baseline. Feature Permutation and Grad-CAM show clearer drops after layer randomisation, which is reflected in decreases in both RRA and DPP. 

Panel~\ref{fig:sanity_pretrained_input} shows the effect of input randomisation. After pixel shuffling removes spatial structure, explanations change and RRA decrease for most methods. The highest decrease is observed for Grad-CAM, where the attribution focused on the cat’s face, despite the misclassification, becomes diffuse hotspots after shuffling. These patterns match the sensitivity measures and indicate that explanations generally depend on image structure. For Grad-CAM can observe the considerable decrease in both RRA and DPP metrics. 

Overall, deeper and especially pretrained models yield explanations that depend more on learned parameters and tend to show greater sensitivity to spatial input structure. This is reflected in higher degradation under layer-wise randomisation and higher sensitivity under pixel permutation. Additional examples are provided in Appendix~\ref{app:examples_sanity_checks}.

\section*{Discussion}
Across datasets and XAI methods, increasing model scale does not yield consistent improvements in explanation quality. In the statistical analysis with Benjamini-Hochberg adjusted $p$-values, many comparisons fail to reject the null hypothesis of equal interpretability scores across scales. When significant differences are observed, the largest models achieve the highest scores in only 12\% of the comparisons, while in the remaining significant cases, smaller models match or exceed their larger counterparts. Moreover, the associated effect sizes are usually negligible to small, which suggests that even statistically significant differences often correspond to only modest practical differences in explanation quality. This finding challenges the assumption that larger models produce more meaningful insights due to their higher representational capacity. It suggests that while deeper models may capture abstract features for classification, these representations do not necessarily translate into better pixel-level attributions. Simpler models often provide explanations that align as well or better with ground truth masks compared to bigger counterparts.

Moreover, the statistical analysis does not show a consistent advantage of pretraining in RRA and DPP. After Benjamini-Hochberg correction, half of the pretraining comparisons are not significant, and among the significant results, improvements are split evenly between pretrained and scratch-trained models. Mean scores are often higher for the ImageNet-pretrained model, but these differences are not systematic enough to translate into a clear overall gain.

The sanity checks provide additional evidence about how explanations depend on model parameters. In scratch-trained models, smaller ResNet and DenseNet variants show slightly lower degradation, which suggests weaker dependence on the trained parameter configuration. The input randomisation across different ResNet and DenseNet model scales show changes that are mostly negligible to small across the sanity checks, indicating limited scale-related variation within these families under this evaluation. For the pretrained Vision Transformers, the sanity checks likewise show mostly negligible to small differences between ViT-B/16 and ViT-L/16, with a medium effect observed only for Saliency on ViT-B/16, which indicates limited variation across scale within this pretrained-only comparison. Pretrained ResNet-50 model show larger degradation under layer-wise weight randomisation with respect to the scratch-trained model, indicating that its attribution maps change more when learned weights are disrupted. Taken together, these findings suggest that pretraining strengthens the link between explanations and learned parameters even when aggregate localisation metrics such as RRA and DPP do not consistently improve.

Results on the Chest X Pneumothorax dataset illustrate a gap between predictive performance and explanation quality that is relevant for medical model evaluation. Models reach reasonable classification performance, yet the corresponding explanations are not aligned with the target regions, and interpretability scores remain near zero. This pattern is consistent with models relying on confounds or imaging artefacts rather than anatomically meaningful cues. These findings support the use of explanation-based metrics for auditing, since evaluation based only on accuracy can miss failures in how predictions are formed.

\subsection*{Influence of Explanation Resolution on Localization Metrics}
We acknowledge that the varying resolutions of the evaluated XAI methods introduce a trade-off when applying pixel-level metrics. Unlike gradient-based methods, which produce fine-grained pixel-level attributions, Grad-CAM generates coarse localization maps by upsampling activation maps from the final convolutional layer. This inherent smoothness may confer an advantage in mass-based evaluations (e.g., DPP) by producing contiguous regions of high relevance that align well with larger ground-truth masks, effectively filtering out the high-frequency gradient noise that penalizes pixel-wise methods. However, our results indicate that this resolution bias does not solely dictate explanation quality. For instance, Feature Permutation also operates at a coarse granularity, assigning importance to $16 \times 16$ pixel patches, yet it yielded distinct performance trends compared to Grad-CAM. Notably, while Grad-CAM performance varied considerably between pretrained and scratch models, Feature Permutation remained relatively stable. This divergence suggests that while upsampling artifacts can influence absolute scores, the relative differences we observed across model scales and pretraining regimes reflect genuine variations in the semantic quality of learned representations rather than simple resolution artifacts.

\subsection*{Implications for Robustness and Reliability}
Robustness to distribution shift and adversarial perturbations is relevant for clinical AI use. Although this study does not include direct robustness evaluations, the interpretability results indicate potential failure modes. Models trained from scratch sometimes fail to localise task-relevant regions despite high classification performance, most clearly in the Pneumothorax dataset, which is consistent with reliance on confounds or dataset artefacts and reduced stability under shifts in acquisition conditions or patient populations.

ImageNet-pretrained models show better alignment with ground truth and higher sensitivity in input-randomisation tests, suggesting stronger dependence on image structure. These observations support using RRA and DPP alongside accuracy when assessing reliability under distribution shift. Future work should test this link directly under controlled out-of-distribution and adversarial settings.

\subsection*{Limitations and Future Work}
This study evaluates explanation quality in binary classification settings. However, multi-class problems may show different interactions between model scale, training regime, and interpretability, particularly when classes share overlapping visual cues.

The evaluation also depends on ground-truth segmentation masks that define regions of interest. While these masks support quantitative localisation metrics, they may not represent the full set of cues used in medical decision-making. The metrics treat overlap with the mask as a proxy for explanation quality, which can be limiting when clinically relevant context lies outside the annotated region. In addition, heatmap-based measures do not cover explanation approaches defined at the level of human concepts or objects. For example, TCAV~\cite{kim2018interpretability} estimates the influence of a concept through concept activation vectors and directional derivatives without producing pixel-level attributions. Extending the analysis to concept-based methods would require concept annotations or object-level intervention protocols in addition to segmentation masks.

Our experiments focus on post-hoc explanations for standard convolutional and transformer architectures. Models designed for interpretability, such as ProtoPNet~\cite{chen2019looks}, provide an alternative that links predictions to learned prototypes and can reduce sensitivity to gradient noise. We do not include such models to maintain a consistent architecture family across scale comparisons. Future work can test how explanation quality in prototype-based and related architectures changes with model size.

Future work could consider architecture-specific training schedules and longer optimization budgets to further examine the robustness of the present findings. In addition, further validation on larger and more diverse datasets would strengthen these findings. The use of greyscale inputs may affect both predictive performance and explanations in non-medical domains where colour carries signal. A practical constraint is that large image datasets rarely provide segmentation masks needed for localisation-based evaluation, as many benchmarks include only class labels. One direction is to use multi-rater datasets such as LIDC-IDRI, where soft masks derived from inter-annotator agreement allow probabilistic assessment of feature importance~\cite{armato2011lung}.

Finally, quantitative measures such as RRA and DPP capture aspects of localisation but do not address whether an explanation supports expert decision-making. In sensitive domains, such as medical image analysis, studies with radiologists that assess usefulness, consistency, and failure modes across explanation methods would provide complementary evidence about clinical relevance.

\section*{Conclusion}
This study examines how model scale relates to the quality of post-hoc visual explanations produced by saliency-style and CAM-style methods. We evaluate eleven architectures (ResNets, DenseNets, and Vision Transformers) across three datasets using five explanation methods. Explanation quality is assessed through spatial alignment with ground-truth masks using Relevance Rank Accuracy (RRA) and the proposed Dual-Polarity Precision (DPP).

Across datasets and methods, larger models do not show a consistent advantage in localisation quality. Most scale comparisons fail to reject the null hypothesis after Benjamini-Hochberg correction, and when significant differences occur, the largest model achieves the highest score in only 12\% of scale comparisons. Smaller architectures, including ResNet-18, DenseNet-121, and ViT-B/16, often match or exceed deeper variants such as ResNet-101, DenseNet-201, and ViT-L/16 as measured by RRA and DPP. Pretraining affects explanations more than scaling in many settings. While improvements in RRA and DPP are not uniform, pretrained models show larger degradation under weight randomisation and higher sensitivity under input randomisation.

The results also show that predictive performance and explanation quality can diverge. On the Pneumothorax task, models reach reasonable classification performance while producing near-zero alignment scores, which indicates that correct predictions can be obtained without localising the annotated regions. This supports evaluating explainability alongside accuracy when selecting models for clinical use.

Overall, model selection should not assume that the largest architecture provides the most reliable explanations. Smaller models can provide similar explanation quality with lower computational cost. Future work should extend the evaluation to multi-class settings and concept-based explanations, and should test whether RRA and DPP relate to reliability under controlled distribution shifts and adversarial perturbations.

\section*{Code availability} The code used in this study is accessible at: \url{https://github.com/mateuszcedro/XAI-at-scale.git}.

\section*{Data availability} The datasets used in this study are available at: \url{https://www.kaggle.com/datasets/anasmohammedtahir/covidqu}, \url{https://www.robots.ox.ac.uk/~vgg/data/pets/}, and \url{https://www.kaggle.com/datasets/vbookshelf/pneumothorax-chest-xray-images-and-masks}.

\section*{Declarations}
Mateusz Cedro received funding from the Flemish Government under the “Onderzoeksprogramma Artifici\"{e}le Intelligentie (AI) Vlaanderen”.

\bibliography{bib}

@misc{thair_COVID-QU-Ex,
    title = {COVID-QU-Ex Dataset},
    url = {https://www.kaggle.com/datasets/anasmohammedtahir/covidqu/},
    author = {Tahir, A.M. and others},
    year = {2021},
    journal = {Kaggle},
    doi = {10.34740/kaggle/dsv/3122958},
    note = {Accessed: 2022-12-01}
}

@article{hedstrom2023quantus,
    author  = {Anna Hedstr{\"o}m and Leander Weber and Daniel Krakowczyk and Dilyara Bareeva and Franz Motzkus and Wojciech Samek and Sebastian Lapuschkin and Marina Marina M.{-}C. H{\"o}hne},
    title   = {Quantus: An Explainable AI Toolkit for Responsible Evaluation of Neural Network Explanations and Beyond},
    journal = {Journal of Machine Learning Research},
    year    = {2023},
    volume  = {24},
    number  = {34},
    pages   = {1--11},
    url     = {http://jmlr.org/papers/v24/22-0142.html}
}

@article{chexpert,
  author       = {Jeremy Irvin and Pranav Rajpurkar and Michael Ko and Yifan Yu and Silviana Ciurea{-}Ilcus and Christopher Chute and Henrik Marklund and Behzad Haghgoo and Robyn L. Ball and Katie S. Shpanskaya and Jayne Seekins and David A. Mong and Safwan S. Halabi and Jesse K. Sandberg and Ricky Jones and David B. Larson and Curtis P. Langlotz and Bhavik N. Patel and Matthew P. Lungren and Andrew Y. Ng},
  title        = {CheXpert: A Large Chest Radiograph Dataset with Uncertainty Labels and Expert Comparison},
  journal      = {CoRR},
  volume       = {abs/1901.07031},
  year         = {2019},
  url          = {http://arxiv.org/abs/1901.07031},
  eprinttype   = {arXiv},
  eprint       = {1901.07031}
}

@article{saporta_benchmarking_2022,
  title     = {Benchmarking saliency methods for chest {X}-ray interpretation},
  volume    = {4},
  issn      = {2522-5839},
  url       = {https://www.nature.com/articles/s42256-022-00536-x},
  doi       = {10.1038/s42256-022-00536-x},
  language  = {en},
  number    = {10},
  urldate   = {2022-12-02},
  journal   = {Nature Machine Intelligence},
  author    = {Saporta, Adriel and Gui, Xiaotong and Agrawal, Ashwin and Pareek, Anuj and Truong, Steven Q. H. and Nguyen, Chanh D. T. and Ngo, Van-Doan and Seekins, Jayne and Blankenberg, Francis G. and Ng, Andrew Y. and Lungren, Matthew P. and Rajpurkar, Pranav},
  month     = oct,
  year      = {2022},
  pages     = {867--878}
}

@article{chowdhury_can_2020,
  title     = {Can {AI} Help in Screening Viral and {COVID}-19 {Pneumonia}?},
  volume    = {8},
  issn      = {2169-3536},
  url       = {https://ieeexplore.ieee.org/document/9144185/},
  doi       = {10.1109/ACCESS.2020.3010287},
  language  = {en},
  urldate   = {2022-12-09},
  journal   = {IEEE Access},
  author    = {Chowdhury, Muhammad E. H. and Rahman, Tawsifur and Khandakar, Amith and Mazhar, Rashid and Kadir, Muhammad Abdul and Mahbub, Zaid Bin and Islam, Khandakar Reajul and Khan, Muhammad Salman and Iqbal, Atif and Emadi, Nasser Al and Reaz, Mamun Bin Ibne and Islam, Mohammad Tariqul},
  year      = {2020},
  pages     = {132665--132676}
}

@article{degerli_covid-19_2021,
  title     = {{COVID}-19 infection map generation and detection from chest {X}-ray images},
  volume    = {9},
  issn      = {2047-2501},
  url       = {https://link.springer.com/10.1007/s13755-021-00146-8},
  doi       = {10.1007/s13755-021-00146-8},
  language  = {en},
  number    = {1},
  urldate   = {2022-12-09},
  journal   = {Health Information Science and Systems},
  author    = {Degerli, Aysen and Ahishali, Mete and Yamac, Mehmet and Kiranyaz, Serkan and Chowdhury, Muhammad E. H. and Hameed, Khalid and Hamid, Tahir and Mazhar, Rashid and Gabbouj, Moncef},
  month     = dec,
  year      = {2021},
  pages     = {15}
}

@article{rahman_exploring_2021,
  title     = {Exploring the effect of image enhancement techniques on {COVID}-19 detection using chest {X}-ray images},
  volume    = {132},
  issn      = {00104825},
  url       = {https://linkinghub.elsevier.com/retrieve/pii/S001048252100113X},
  doi       = {10.1016/j.compbiomed.2021.104319},
  language  = {en},
  urldate   = {2022-12-09},
  journal   = {Computers in Biology and Medicine},
  author    = {Rahman, Tawsifur and Khandakar, Amith and Qiblawey, Yazan and Tahir, Anas and Kiranyaz, Serkan and Abul Kashem, Saad Bin and Islam, Mohammad Tariqul and Al Maadeed, Somaya and Zughaier, Susu M. and Khan, Muhammad Salman and Chowdhury, Muhammad E.H.},
  month     = may,
  year      = {2021},
  pages     = {104319}
}

@inproceedings{davenport_potential_2019,
  title     = {The potential for artificial intelligence in healthcare},
  language  = {en},
  author    = {Davenport, Thomas and Kalakota, Ravi},
  year      = {2019},
}

@article{lecun_deep_2015,
	title = {Deep learning},
	volume = {521},
	issn = {0028-0836, 1476-4687},
	url = {https://www.nature.com/articles/nature14539},
	doi = {10.1038/nature14539},
	language = {en},
	number = {7553},
	urldate = {2023-11-08},
	journal = {Nature},
	author = {LeCun, Yann and Bengio, Yoshua and Hinton, Geoffrey},
	month = may,
	year = {2015},
	pages = {436--444},
}

@inproceedings{cabitza_unintended_2017,
	title = {Unintended {Consequences} of {Machine} {Learning} in {Medicine}},
	language = {en},
	author = {Cabitza, Federico and Rasoini, Raffaele and Gensini, Gian Franco},
	year = {2017},
}

@article{hinton_deep_2018,
	title = {Deep {Learning}—{A} {Technology} {With} the {Potential} to {Transform} {Health} {Care}},
	volume = {320},
	issn = {0098-7484},
	url = {http://jama.jamanetwork.com/article.aspx?doi=10.1001/jama.2018.11100},
	doi = {10.1001/jama.2018.11100},
	language = {en},
	number = {11},
	urldate = {2023-11-08},
	journal = {JAMA},
	author = {Hinton, Geoffrey},
	month = sep,
	year = {2018},
	pages = {1101},
}

@misc{zhaounderstanding,
  title={Understanding scene in the wild},
  author={Zhao, Hengshuang and Shi, Jianping and Qi, Xiaojuan and Wang, Xiaogang and Xiao, Tong and Jia, Jiaya},
    year = {2016}
}

@article{brown_language_2020,
  author       = {Tom B. Brown and others},
  title        = {Language Models are Few-Shot Learners},
  journal      = {CoRR},
  volume       = {abs/2005.14165},
  year         = {2020},
  url          = {https://arxiv.org/abs/2005.14165},
  eprinttype    = {arXiv},
  eprint       = {2005.14165},
  bibsource    = {dblp computer science bibliography, https://dblp.org}
}

@article{sarwinda_deep_2021,
  title     = {Deep Learning in Image Classification using Residual Network (ResNet) Variants for Detection of Colorectal Cancer},
  volume    = {179},
  issn      = {1877-0509},
  url       = {https://linkinghub.elsevier.com/retrieve/pii/S1877050921000284},
  doi       = {10.1016/j.procs.2021.01.025},
  language  = {en},
  urldate   = {2023-11-07},
  journal   = {Procedia Computer Science},
  author    = {Sarwinda, Devvi and Paradisa, Radifa Hilya and Bustamam, Alhadi and Anggia, Pinkie},
  year      = {2021},
  pages     = {423--431}
}

@misc{eigen_understanding_2014,
	title = {Understanding {Deep} {Architectures} using a {Recursive} {Convolutional} {Network}},
	url = {http://arxiv.org/abs/1312.1847},
	language = {en},
	urldate = {2023-11-01},
	publisher = {arXiv},
	author = {Eigen, David and Rolfe, Jason and Fergus, Rob and LeCun, Yann},
	month = feb,
	year = {2014},
	note = {arXiv:1312.1847 [cs]},
}

@inproceedings{khan_evaluating_2018,
	address = {Chengdu China},
	title = {Evaluating the {Performance} of {ResNet} {Model} {Based} on {Image} {Recognition}},
	isbn = {978-1-4503-6419-5},
	url = {https://dl.acm.org/doi/10.1145/3194452.3194461},
	doi = {10.1145/3194452.3194461},
	language = {en},
	urldate = {2023-11-01},
	booktitle = {Proceedings of the 2018 {International} {Conference} on {Computing} and {Artificial} {Intelligence}},
	publisher = {ACM},
	author = {Khan, Riaz Ullah and Zhang, Xiaosong and Kumar, Rajesh and Aboagye, Emelia Opoku},
	month = mar,
	year = {2018},
	pages = {86--90},
}

@misc{wu_wider_2016,
	title = {Wider or {Deeper}: {Revisiting} the {ResNet} {Model} for {Visual} {Recognition}},
	shorttitle = {Wider or {Deeper}},
	url = {http://arxiv.org/abs/1611.10080},
	language = {en},
	urldate = {2023-11-02},
	publisher = {arXiv},
	author = {Wu, Zifeng and Shen, Chunhua and Hengel, Anton van den},
	month = nov,
	year = {2016},
	note = {arXiv:1611.10080 [cs]},
}

@article{menghani_efficient_2023,
	title = {Efficient {Deep} {Learning}: {A} {Survey} on {Making} {Deep} {Learning} {Models} {Smaller}, {Faster}, and {Better}},
	volume = {55},
	issn = {0360-0300, 1557-7341},
	shorttitle = {Efficient {Deep} {Learning}},
	url = {https://dl.acm.org/doi/10.1145/3578938},
	doi = {10.1145/3578938},
	language = {en},
	number = {12},
	urldate = {2023-11-02},
	journal = {ACM Computing Surveys},
	author = {Menghani, Gaurav},
	month = dec,
	year = {2023},
	pages = {1--37},
}

@inproceedings{guo_classification_2019,
	address = {Xiamen, China},
	title = {Classification of {Thyroid} {Ultrasound} {Standard} {Plane} {Images} using {ResNet}-18 {Networks}},
	isbn = {978-1-72812-458-2},
	url = {https://ieeexplore.ieee.org/document/8925267/},
	doi = {10.1109/ICASID.2019.8925267},
	language = {en},
	urldate = {2023-11-02},
	booktitle = {2019 {IEEE} 13th {International} {Conference} on {Anti}-counterfeiting, {Security}, and {Identification} ({ASID})},
	publisher = {IEEE},
	author = {Guo, Minghui and Du, Yongzhao},
	month = oct,
	year = {2019},
	pages = {324--328},
}

@article{showkat_efficacy_2022,
	title = {Efficacy of {Transfer} {Learning}-based {ResNet} models in {Chest} {X}-ray image classification for detecting {COVID}-19 {Pneumonia}},
	volume = {224},
	issn = {01697439},
	url = {https://linkinghub.elsevier.com/retrieve/pii/S0169743922000454},
	doi = {10.1016/j.chemolab.2022.104534},
	language = {en},
	urldate = {2023-11-02},
	journal = {Chemometrics and Intelligent Laboratory Systems},
	author = {Showkat, Sadia and Qureshi, Shaima},
	month = may,
	year = {2022},
	pages = {104534},
}

@inproceedings{molnar_interpretable_2020,
  title     = {Interpretable Machine Learning -- A Brief History, State-of-the-Art and Challenges},
  volume    = {1323},
  url       = {http://arxiv.org/abs/2010.09337},
  language  = {en},
  urldate   = {2023-11-03},
  author    = {Molnar, Christoph and Casalicchio, Giuseppe and Bischl, Bernd},
  year      = {2020},
  doi       = {10.1007/978-3-030-65965-3_28},
  note      = {arXiv:2010.09337 [cs, stat]},
  pages     = {417--431}
}

@misc{ribeiro_why_2016,
	title = {"{Why} {Should} {I} {Trust} {You}?": {Explaining} the {Predictions} of {Any} {Classifier}},
	shorttitle = {"{Why} {Should} {I} {Trust} {You}?},
	url = {http://arxiv.org/abs/1602.04938},
	language = {en},
	urldate = {2023-11-03},
	publisher = {arXiv},
	author = {Ribeiro, Marco Tulio and Singh, Sameer and Guestrin, Carlos},
	month = aug,
	year = {2016},
	note = {arXiv:1602.04938 [cs, stat]},
}

@article{guidotti_survey_2019,
	title = {A {Survey} of {Methods} for {Explaining} {Black} {Box} {Models}},
	volume = {51},
	issn = {0360-0300, 1557-7341},
	url = {https://dl.acm.org/doi/10.1145/3236009},
	doi = {10.1145/3236009},
	language = {en},
	number = {5},
	urldate = {2023-11-03},
	journal = {ACM Computing Surveys},
	author = {Guidotti, Riccardo and Monreale, Anna and Ruggieri, Salvatore and Turini, Franco and Giannotti, Fosca and Pedreschi, Dino},
	month = sep,
	year = {2019},
	pages = {1--42},
}

@inproceedings{katuwal_machine_nodate,
	title = {Machine {Learning} {Model} {Interpretability} for {Precision} {Medicine}},
	language = {en},
	author = {Katuwal, Gajendra J and Chen, Robert},
    year = {2016},
}

@inproceedings{che_interpretable_nodate,
	title = {Interpretable {Deep} {Models} for {ICU} {Outcome} {Prediction}},
	language = {en},
	author = {Che, Zhengping and Purushotham, Sanjay and Khemani, Robinder and Liu, Yan},
    year = {2017},
}

@article{secinaro_role_2021,
	title = {The role of artificial intelligence in healthcare: a structured literature review},
	volume = {21},
	issn = {1472-6947},
	shorttitle = {The role of artificial intelligence in healthcare},
	url = {https://bmcmedinformdecismak.biomedcentral.com/articles/10.1186/s12911-021-01488-9},
	doi = {10.1186/s12911-021-01488-9},
	language = {en},
	number = {1},
	urldate = {2023-11-04},
	journal = {BMC Medical Informatics and Decision Making},
	author = {Secinaro, Silvana and Calandra, Davide and Secinaro, Aurelio and Muthurangu, Vivek and Biancone, Paolo},
	month = dec,
	year = {2021},
	pages = {125},
}

@book{samek_explainable_2019,
	address = {Cham},
	series = {Lecture {Notes} in {Computer} {Science}},
	title = {Explainable {AI}: {Interpreting}, {Explaining} and {Visualizing} {Deep} {Learning}},
	volume = {11700},
	isbn = {978-3-030-28953-9 978-3-030-28954-6},
	shorttitle = {Explainable {AI}},
	url = {http://link.springer.com/10.1007/978-3-030-28954-6},
	language = {en},
	urldate = {2023-11-05},
	publisher = {Springer International Publishing},
	editor = {Samek, Wojciech and Montavon, Grégoire and Vedaldi, Andrea and Hansen, Lars Kai and Müller, Klaus-Robert},
	year = {2019},
	doi = {10.1007/978-3-030-28954-6},
}

@misc{smilkov_smoothgrad_2017,
	title = {{SmoothGrad}: removing noise by adding noise},
	shorttitle = {{SmoothGrad}},
	url = {http://arxiv.org/abs/1706.03825},
	language = {en},
	urldate = {2023-11-05},
	publisher = {arXiv},
	author = {Smilkov, Daniel and Thorat, Nikhil and Kim, Been and Viégas, Fernanda and Wattenberg, Martin},
	month = jun,
	year = {2017},
	note = {arXiv:1706.03825 [cs, stat]},
}

@article{bach_pixel-wise_2015,
	title = {On {Pixel}-{Wise} {Explanations} for {Non}-{Linear} {Classifier} {Decisions} by {Layer}-{Wise} {Relevance} {Propagation}},
	volume = {10},
	issn = {1932-6203},
	url = {https://dx.plos.org/10.1371/journal.pone.0130140},
	doi = {10.1371/journal.pone.0130140},
	language = {en},
	number = {7},
	urldate = {2023-11-05},
	journal = {PLOS ONE},
	author = {Bach, Sebastian and Binder, Alexander and Montavon, Grégoire and Klauschen, Frederick and Müller, Klaus-Robert and Samek, Wojciech},
	editor = {Suarez, Oscar Deniz},
	month = jul,
	year = {2015},
	pages = {e0130140},
}

@misc{shrikumar_learning_2019,
	title = {Learning {Important} {Features} {Through} {Propagating} {Activation} {Differences}},
	url = {http://arxiv.org/abs/1704.02685},
	language = {en},
	urldate = {2023-11-05},
	publisher = {arXiv},
	author = {Shrikumar, Avanti and Greenside, Peyton and Kundaje, Anshul},
	month = oct,
	year = {2019},
	note = {arXiv:1704.02685 [cs]},
}

@article{apley_visualizing_2020,
	title = {Visualizing the {Effects} of {Predictor} {Variables} in {Black} {Box} {Supervised} {Learning} {Models}},
	volume = {82},
	issn = {1369-7412, 1467-9868},
	url = {https://academic.oup.com/jrsssb/article/82/4/1059/7056085},
	doi = {10.1111/rssb.12377},
	language = {en},
	number = {4},
	urldate = {2023-11-05},
	journal = {Journal of the Royal Statistical Society Series B: Statistical Methodology},
	author = {Apley, Daniel W. and Zhu, Jingyu},
	month = sep,
	year = {2020},
	pages = {1059--1086},
}

@article{rajkomar_machine_2019,
	title = {Machine {Learning} in {Medicine}},
	volume = {380},
	issn = {0028-4793, 1533-4406},
	url = {http://www.nejm.org/doi/10.1056/NEJMra1814259},
	doi = {10.1056/NEJMra1814259},
	language = {en},
	number = {14},
	urldate = {2023-11-05},
	journal = {New England Journal of Medicine},
	author = {Rajkomar, Alvin and Dean, Jeffrey and Kohane, Isaac},
	month = apr,
	year = {2019},
	pages = {1347--1358},
}

@article{holzinger_causability_2019,
	title = {Causability and explainability of artificial intelligence in medicine},
	volume = {9},
	issn = {1942-4787, 1942-4795},
	url = {https://wires.onlinelibrary.wiley.com/doi/10.1002/widm.1312},
	doi = {10.1002/widm.1312},
	language = {en},
	number = {4},
	urldate = {2023-11-05},
	journal = {WIREs Data Mining and Knowledge Discovery},
	author = {Holzinger, Andreas and Langs, Georg and Denk, Helmut and Zatloukal, Kurt and Müller, Heimo},
	month = jul,
	year = {2019},
	pages = {e1312},
}

@misc{doshi-velez_towards_2017,
	title = {Towards {A} {Rigorous} {Science} of {Interpretable} {Machine} {Learning}},
	url = {http://arxiv.org/abs/1702.08608},
	language = {en},
	urldate = {2023-11-05},
	publisher = {arXiv},
	author = {Doshi-Velez, Finale and Kim, Been},
	month = mar,
	year = {2017},
	note = {arXiv:1702.08608},
}

@misc{hooker_benchmark_2019,
	title = {A {Benchmark} for {Interpretability} {Methods} in {Deep} {Neural} {Networks}},
	url = {http://arxiv.org/abs/1806.10758},
	language = {en},
	urldate = {2023-11-05},
	publisher = {arXiv},
	author = {Hooker, Sara and Erhan, Dumitru and Kindermans, Pieter-Jan and Kim, Been},
	month = nov,
	year = {2019},
	note = {arXiv:1806.10758 [cs, stat]},
}

@misc{bykov_noisegrad_2022,
	title = {{NoiseGrad}: {Enhancing} {Explanations} by {Introducing} {Stochasticity} to {Model} {Weights}},
	shorttitle = {{NoiseGrad}},
	url = {http://arxiv.org/abs/2106.10185},
	language = {en},
	urldate = {2023-11-05},
	publisher = {arXiv},
	author = {Bykov, Kirill and Hedström, Anna and Nakajima, Shinichi and Höhne, Marina M.-C.},
	month = may,
	year = {2022},
	note = {arXiv:2106.10185 [cs]},
}

@article{arras_ground_2022,
	title = {Ground {Truth} {Evaluation} of {Neural} {Network} {Explanations} with {CLEVR}-{XAI}},
	volume = {81},
	issn = {15662535},
	url = {http://arxiv.org/abs/2003.07258},
	doi = {10.1016/j.inffus.2021.11.008},
	language = {en},
	urldate = {2023-11-05},
	journal = {Information Fusion},
	author = {Arras, Leila and Osman, Ahmed and Samek, Wojciech},
	month = may,
	year = {2022},
	note = {arXiv:2003.07258 [cs, eess]},
	pages = {14--40},
}

@misc{adebayo_sanity_2020,
	title = {Sanity {Checks} for {Saliency} {Maps}},
	url = {http://arxiv.org/abs/1810.03292},
	language = {en},
	urldate = {2023-11-06},
	publisher = {arXiv},
	author = {Adebayo, Julius and Gilmer, Justin and Muelly, Michael and Goodfellow, Ian and Hardt, Moritz and Kim, Been},
	month = nov,
	year = {2020},
	note = {arXiv:1810.03292 [cs, stat]},
}

@misc{he_deep_2015,
	title = {Deep {Residual} {Learning} for {Image} {Recognition}},
	url = {http://arxiv.org/abs/1512.03385},
	language = {en},
	urldate = {2023-10-11},
	publisher = {arXiv},
	author = {He, Kaiming and Zhang, Xiangyu and Ren, Shaoqing and Sun, Jian},
	month = dec,
	year = {2015},
	note = {arXiv:1512.03385 [cs]},
}

@misc{lundberg_unified_2017,
	title = {A {Unified} {Approach} to {Interpreting} {Model} {Predictions}},
	url = {http://arxiv.org/abs/1705.07874},
	language = {en},
	urldate = {2023-10-11},
	publisher = {arXiv},
	author = {Lundberg, Scott and Lee, Su-In},
	month = nov,
	year = {2017},
	note = {arXiv:1705.07874 [cs, stat]},
}

@inproceedings{sundararajan_axiomatic_2017,
  title={Axiomatic attribution for deep networks},
  author={Sundararajan, Mukund and Taly, Ankur and Yan, Qiqi},
  booktitle={International conference on machine learning},
  pages={3319--3328},
  year={2017},
  organization={PMLR}
}

@inproceedings{morch_visualization_1995,
  address   = {Perth, WA, Australia},
  title     = {Visualization of neural networks using saliency maps},
  volume    = {4},
  isbn      = {978-0-7803-2768-9},
  url       = {http://ieeexplore.ieee.org/document/488997/},
  doi       = {10.1109/ICNN.1995.488997},
  language  = {en},
  urldate   = {2023-10-11},
  booktitle = {Proceedings of ICNN'95 - International Conference on Neural Networks},
  publisher = {IEEE},
  author    = {Morch, N. J. S. and Kjems, U. and Hansen, L. K. and Svarer, C. and Law, I. and Lautrup, B. and Strother, S. and Rehm, K.},
  year      = {1995},
  pages     = {2085--2090}
}

@inproceedings{baehrens_how_nodate,
  title     = {How to Explain Individual Classification Decisions},
  language  = {en},
  year      = {2010},
  author    = {Baehrens, David and Schroeter, Timon and Harmeling, Stefan and Kawanabe, Motoaki and Hansen, Katja},
  note      = {arXiv preprint}
}

@book{Shapley_1953,
    author = {Lloyd S Shapley},
    title = {Contributions to the Theory of Games},
    year = {1953},
    note = {“A value for n-person games”, pp. 307–317.}
}

@misc{simonyan_deep_2014,
	title = {Deep {Inside} {Convolutional} {Networks}: {Visualising} {Image} {Classification} {Models} and {Saliency} {Maps}},
	shorttitle = {Deep {Inside} {Convolutional} {Networks}},
	url = {http://arxiv.org/abs/1312.6034},
	language = {en},
	urldate = {2023-10-23},
	publisher = {arXiv},
	author = {Simonyan and others},
	month = apr,
	year = {2013},
	note = {arXiv:1312.6034 [cs]},
}

@book{molnar2022,
  title      = {Interpretable Machine Learning},
  author     = {Christoph Molnar},
  year       = {2022},
  subtitle   = {A Guide for Making Black Box Models Explainable},
  edition    = {2},
  url        = {https://christophm.github.io/interpretable-ml-book}
}

@InProceedings{imagenet,
  title={Imagenet: A large-scale hierarchical image database},
  author={Deng, Jia and Dong, Wei and Socher, Richard and Li, Li-Jia and Li, Kai and Fei-Fei, Li},
  booktitle={2009 IEEE conference on computer vision and pattern recognition},
  pages={248--255},
  year={2009},
  organization={Ieee}
}

@InProceedings{densely,
  title={Densely connected convolutional networks},
  author={Huang, Gao and Liu, Zhuang and Van Der Maaten, Laurens and Weinberger, Kilian Q},
  booktitle={Proceedings of the IEEE conference on computer vision and pattern recognition},
  pages={4700--4708},
  year={2017}
}

@InProceedings{cedro2024graphxain,
author="Cedro, Mateusz
and Martens, David",
editor="Guidotti, Riccardo
and Schmid, Ute
and Longo, Luca",
title="{G}raph{XAIN}: {N}arratives to {E}xplain {G}raph {N}eural {N}etworks",
booktitle="Explainable Artificial Intelligence",
year="2026",
publisher="Springer Nature Switzerland",
address="Cham",
pages="91--114",
isbn="978-3-032-08327-2"
}

@article{Dobrzycka31122025,
author = {Margarita Dobrzycka and Anetta Sulewska and Joanna Konopinska and Piotr Karabowicz and Angelika Charkiewicz and Kinga Golaszewska and Mateusz Cedro and Przemysław Biecek and Jacek Niklinski and Alicja Charkiewicz and Radosław Charkiewicz},
title = {Machine learning-based identification of small RNA signatures in aqueous humor as a step toward precision diagnosis of glaucoma},
journal = {Annals of Medicine},
volume = {57},
number = {1},
pages = {2568119},
year = {2025},
publisher = {Taylor \& Francis},
doi = {10.1080/07853890.2025.2568119},
    note ={PMID: 41047921},
URL = {https://doi.org/10.1080/07853890.2025.2568119},
eprint = {https://doi.org/10.1080/07853890.2025.2568119}
}

@inproceedings{selvaraju2017grad,
  title={Grad-cam: Visual explanations from deep networks via gradient-based localization},
  author={Selvaraju, Ramprasaath R and Cogswell, Michael and Das, Abhishek and Vedantam, Ramakrishna and Parikh, Devi and Batra, Dhruv},
  booktitle={Proceedings of the IEEE international conference on computer vision},
  pages={618--626},
  year={2017}
}

@inproceedings{huang2017densely,
  title={Densely connected convolutional networks},
  author={Huang, Gao and Liu, Zhuang and Van Der Maaten, Laurens and Weinberger, Kilian Q},
  booktitle={Proceedings of the IEEE conference on computer vision and pattern recognition},
  pages={4700--4708},
  year={2017}
}

@inproceedings{irvin2019chexpert,
  title={Chexpert: A large chest radiograph dataset with uncertainty labels and expert comparison},
  author={Irvin, Jeremy and Rajpurkar, Pranav and Ko, Michael and Yu, Yifan and Ciurea-Ilcus, Silviana and Chute, Chris and Marklund, Henrik and Haghgoo, Behzad and Ball, Robyn and Shpanskaya, Katie and others},
  booktitle={Proceedings of the AAAI conference on artificial intelligence},
  volume={33},
  number={01},
  pages={590--597},
  year={2019}
}

@article{rajpurkar2017chexnet,
  title={Chexnet: Radiologist-level pneumonia detection on chest x-rays with deep learning},
  author={Rajpurkar, Pranav and Irvin, Jeremy and Zhu, Kaylie and Yang, Brandon and Mehta, Hershel and Duan, Tony and Ding, Daisy and Bagul, Aarti and Langlotz, Curtis and Shpanskaya, Katie and others},
  journal={arXiv preprint arXiv:1711.05225},
  year={2017}
}

@article{fisher2019all,
  title={All models are wrong, but many are useful: Learning a variable's importance by studying an entire class of prediction models simultaneously},
  author={Fisher, Aaron and Rudin, Cynthia and Dominici, Francesca},
  journal={Journal of Machine Learning Research},
  volume={20},
  number={177},
  pages={1--81},
  year={2019}
}

@article{breiman2001random,
  title={Random forests},
  author={Breiman, Leo},
  journal={Machine learning},
  volume={45},
  number={1},
  pages={5--32},
  year={2001},
  publisher={Springer}
}

@article{miller2019explanation,
  title={Explanation in artificial intelligence: Insights from the social sciences},
  author={Miller, Tim},
  journal={Artificial intelligence},
  volume={267},
  pages={1--38},
  year={2019},
  publisher={Elsevier}
}

@article{hestness2017deep,
  title={Deep learning scaling is predictable, empirically},
  author={Hestness, Joel and Narang, Sharan and Ardalani, Newsha and Diamos, Gregory and Jun, Heewoo and Kianinejad, Hassan and Patwary, Md Mostofa Ali and Yang, Yang and Zhou, Yanqi},
  journal={arXiv preprint arXiv:1712.00409},
  year={2017}
}

@article{dosovitskiy2020image,
  title={An image is worth 16x16 words: Transformers for image recognition at scale},
  author={Dosovitskiy, Alexey},
  journal={arXiv preprint arXiv:2010.11929},
  year={2020},
}

@article{bahri2024explaining,
  title={Explaining neural scaling laws},
  author={Bahri, Yasaman and Dyer, Ethan and Kaplan, Jared and Lee, Jaehoon and Sharma, Utkarsh},
  journal={Proceedings of the National Academy of Sciences},
  volume={121},
  number={27},
  pages={e2311878121},
  year={2024},
  publisher={National Academy of Sciences}
}

@inproceedings{brigato2021close,
  title={A close look at deep learning with small data},
  author={Brigato, Lorenzo and Iocchi, Luca},
  booktitle={2020 25th international conference on pattern recognition (ICPR)},
  pages={2490--2497},
  year={2021},
  organization={IEEE}
}

@inproceedings{balogh2023evaluation,
  title={Evaluation of explainable ai localisation performance using relevance f-score},
  author={Balogh, Gregory and McLaughlin, Niall and Rainer, Austen},
  booktitle={25th Irish Machine Vision and Image Processing Conference 2023},
  pages={96--103},
  year={2023},
  organization={Irish Pattern Recognition \& Classification Society},
}

@inproceedings{kim2018interpretability,
  title={Interpretability beyond feature attribution: Quantitative testing with concept activation vectors (tcav)},
  author={Kim, Been and Wattenberg, Martin and Gilmer, Justin and Cai, Carrie and Wexler, James and Viegas, Fernanda and others},
  booktitle={International conference on machine learning},
  pages={2668--2677},
  year={2018},
  organization={PMLR}
}

@article{chen2019looks,
  title={This looks like that: deep learning for interpretable image recognition},
  author={Chen, Chaofan and Li, Oscar and Tao, Daniel and Barnett, Alina and Rudin, Cynthia and Su, Jonathan K},
  journal={Advances in neural information processing systems},
  volume={32},
  year={2019}
}

@article{armato2011lung,
  title={The lung image database consortium (LIDC) and image database resource initiative (IDRI): a completed reference database of lung nodules on CT scans},
  author={Armato III, Samuel G and McLennan, Geoffrey and Bidaut, Luc and McNitt-Gray, Michael F and Meyer, Charles R and Reeves, Anthony P and Zhao, Binsheng and Aberle, Denise R and Henschke, Claudia I and Hoffman, Eric A and others},
  journal={Medical physics},
  volume={38},
  number={2},
  pages={915--931},
  year={2011},
  publisher={Wiley Online Library}
}

@inproceedings{chefer2021transformer,
  title={Transformer interpretability beyond attention visualization},
  author={Chefer, Hila and Gur, Shir and Wolf, Lior},
  booktitle={Proceedings of the IEEE/CVF conference on computer vision and pattern recognition},
  pages={782--791},
  year={2021}
}

@article{kokhlikyan2020captum,
  title={Captum: A unified and generic model interpretability library for pytorch},
  author={Kokhlikyan, Narine and Miglani, Vivek and Martin, Miguel and Wang, Edward and Alsallakh, Bilal and Reynolds, Jonathan and Melnikov, Alexander and Kliushkina, Natalia and Araya, Carlos and Yan, Siqi and others},
  journal={arXiv preprint arXiv:2009.07896},
  year={2020}
}

\appendix
\section{Statistical Significance Tests}
\label{app:stat}
This section describes the statistical procedures used to compare explanation quality across model variants. We use non-parametric tests because the evaluation metrics do not follow a normal distribution. For comparisons with more than two groups in the ResNet and DenseNet scale analyses, we apply the Kruskal--Wallis H-test. For pairwise comparisons in the Vision Transformer scale analysis and in the pretraining analyses, we apply the Mann--Whitney U-test. To control the false discovery rate across the set of hypothesis tests spanning datasets, models, and XAI methods, we adjust all $p$-values using the Benjamini--Hochberg procedure. Results are computed from explanations obtained for 200 randomly sampled test images per seed across three seeds, yielding 600 samples per group.

We report effect sizes alongside significance tests to describe the magnitude of observed differences. For Kruskal--Wallis tests, we report $\eta^2$. For Mann--Whitney tests, we report Cliff's $\delta$. We also estimate post-hoc statistical power using bootstrap resampling. For each comparison, we resample the observed group distributions with replacement 1{,}000 times, re-run the corresponding test, and define power as the proportion of resamples with $p<0.05$. Table~\ref{tab:stats_all} reports the full set of test outcomes, including Benjamini--Hochberg adjusted $p$-values, effect sizes with interpretations, and power estimates.

\begin{table}[ht!]
\centering
\caption{Statistical comparison of XAI methods across model scales and pretraining for Relevance Rank Accuracy and Dual-Polarity Precision scores. $p$-values are Benjamini-Hochberg adjusted. Effect sizes are $\eta^2$ (Kruskal-Wallis) or $\delta$ (Mann-Whitney U). We estimate post-hoc power (1-$\beta$) by bootstrap resampling using unadjusted $p$-values. $^{*}$ indicates significance at $\alpha = 0.05$ after correction. Note: \textsuperscript{$\dagger$} indicates CheXpert-pretrained DenseNet-121 model that did not converge within the training budget of 100 epochs across all seeds.}

\label{tab:stats_all}

\begin{subtable}[t]{\textwidth}
\centering
\small
\vspace{-1.35em}
\caption{COVID-QU-Ex Dataset}
\label{tab:stats_all_a}
\vspace{-0.8em}

\begin{tabular}{lllcccc}
\toprule
Group & Metric & Method & $p$-value (BH-adj.) & Effect Size & Interp. & Power \\
\midrule
\multirow{10}{*}{ResNet Scale} & \multirow{5}{*}{Relevance Rank Accuracy} & Saliency & \adjp{0.42}{\phantom{*}} & $\eta^2$=0.00 & Negligible & 0.28 \\
 &  & Integrated Gradients & \adjp{0.04}{*} & $\eta^2$=0.00 & Negligible & 0.72 \\
 &  & GradientSHAP & \adjp{0.16}{\phantom{*}} & $\eta^2$=0.00 & Negligible & 0.47 \\
 &  & Grad-CAM & \adjp{<0.01}{*} & $\eta^2$=0.01 & Small & 0.99 \\
 &  & Feature Permutation & \adjp{<0.01}{*} & $\eta^2$=0.03 & Small & 1.00 \\
\cmidrule(lr){2-7}
 & \multirow{5}{*}{Dual Polarity Precision} & Saliency & \adjp{<0.01}{*} & $\eta^2$=0.02 & Small & 1.00 \\
 &  & Integrated Gradients & \adjp{<0.01}{*} & $\eta^2$=0.02 & Small & 1.00 \\
 &  & GradientSHAP & \adjp{<0.01}{*} & $\eta^2$=0.01 & Small & 1.00 \\
 &  & Grad-CAM & \adjp{0.03}{*} & $\eta^2$=0.00 & Negligible & 0.75 \\
 &  & Feature Permutation & \adjp{<0.01}{*} & $\eta^2$=0.02 & Small & 1.00 \\
\midrule

\multirow{10}{*}{DenseNet Scale} & \multirow{5}{*}{Relevance Rank Accuracy} & Saliency & \adjp{<0.01}{*} & $\eta^2$=0.02 & Small & 1.00 \\
 &  & Integrated Gradients & \adjp{0.04}{*} & $\eta^2$=0.00 & Negligible & 0.68 \\
 &  & GradientSHAP & \adjp{0.03}{*} & $\eta^2$=0.00 & Negligible & 0.74 \\
 &  & Grad-CAM & \adjp{<0.01}{*} & $\eta^2$=0.05 & Small & 1.00 \\
 &  & Feature Permutation & \adjp{<0.01}{*} & $\eta^2$=0.01 & Small & 0.97 \\
\cmidrule(lr){2-7}
 & \multirow{5}{*}{Dual Polarity Precision} & Saliency & \adjp{<0.01}{*} & $\eta^2$=0.01 & Small & 0.99 \\
 &  & Integrated Gradients & \adjp{<0.01}{*} & $\eta^2$=0.01 & Small & 1.00 \\
 &  & GradientSHAP & \adjp{<0.01}{*} & $\eta^2$=0.01 & Small & 0.94 \\
 &  & Grad-CAM & \adjp{<0.01}{*} & $\eta^2$=0.03 & Small & 1.00 \\
 &  & Feature Permutation & \adjp{0.13}{\phantom{*}} & $\eta^2$=0.00 & Negligible & 0.45 \\
\midrule

\multirow{10}{*}{ViT (pretrained) Scale} & \multirow{5}{*}{Relevance Rank Accuracy} & Saliency & \adjp{0.03}{*} & $\delta$=0.08 & Negligible & 0.71 \\
 &  & Integrated Gradients & \adjp{0.81}{\phantom{*}} & $\delta$=-0.01 & Negligible & 0.06 \\
 &  & GradientSHAP & \adjp{0.80}{\phantom{*}} & $\delta$=0.01 & Negligible & 0.05 \\
 &  & Grad-CAM & \adjp{0.24}{\phantom{*}} & $\delta$=-0.04 & Negligible & 0.23 \\
 &  & Feature Permutation & \adjp{0.09}{\phantom{*}} & $\delta$=0.06 & Negligible & 0.46 \\
\cmidrule(lr){2-7}
 & \multirow{5}{*}{Dual Polarity Precision} & Saliency & \adjp{<0.01}{*} & $\delta$=-0.10 & Negligible & 0.85 \\
 &  & Integrated Gradients & \adjp{<0.01}{*} & $\delta$=-0.16 & Small & 1.00 \\
 &  & GradientSHAP & \adjp{<0.01}{*} & $\delta$=-0.15 & Small & 0.99 \\
 &  & Grad-CAM & \adjp{0.18}{\phantom{*}} & $\delta$=-0.05 & Negligible & 0.30 \\
 &  & Feature Permutation & \adjp{0.78}{\phantom{*}} & $\delta$=-0.01 & Negligible & 0.07 \\
 \midrule

\multirow{10}{*}{ResNet Pretraining} & \multirow{5}{*}{Relevance Rank Accuracy} & Saliency & \adjp{<0.01}{*} & $\delta$=0.25 & Small & 1.00 \\
 &  & Integrated Gradients & \adjp{0.07}{\phantom{*}} & $\delta$=0.07 & Negligible & 0.49 \\
 &  & GradientSHAP & \adjp{0.02}{*} & $\delta$=0.09 & Negligible & 0.74 \\
 &  & Grad-CAM & \adjp{<0.01}{*} & $\delta$=-0.14 & Negligible & 0.99 \\
 &  & Feature Permutation & \adjp{0.13}{\phantom{*}} & $\delta$=-0.05 & Negligible & 0.38 \\
\cmidrule(lr){2-7}
 & \multirow{5}{*}{Dual Polarity Precision} & Saliency & \adjp{<0.01}{*} & $\delta$=-0.16 & Small & 0.99 \\
 &  & Integrated Gradients & \adjp{<0.01}{*} & $\delta$=-0.12 & Negligible & 0.94 \\
 &  & GradientSHAP & \adjp{0.30}{\phantom{*}} & $\delta$=-0.04 & Negligible & 0.23 \\
 &  & Grad-CAM & \adjp{0.03}{*} & $\delta$=-0.08 & Negligible & 0.70 \\
 &  & Feature Permutation & \adjp{0.26}{\phantom{*}} & $\delta$=-0.04 & Negligible & 0.22 \\
\midrule

\multirow{10}{*}{DenseNet Pretraining\textsuperscript{$\dagger$}} & \multirow{5}{*}{Relevance Rank Accuracy} & Saliency & \adjp{0.05}{\phantom{*}} & $\delta$=-0.07 & Negligible & 0.57 \\
 &  & Integrated Gradients & \adjp{0.03}{*} & $\delta$=-0.08 & Negligible & 0.69 \\
 &  & GradientSHAP & \adjp{0.03}{*} & $\delta$=-0.08 & Negligible & 0.69 \\
 &  & Grad-CAM & \adjp{<0.01}{*} & $\delta$=-0.34 & Medium & 1.00 \\
 &  & Feature Permutation & \adjp{<0.01}{*} & $\delta$=0.22 & Small & 1.00 \\
\cmidrule(lr){2-7}
 & \multirow{5}{*}{Dual Polarity Precision} & Saliency & \adjp{0.12}{\phantom{*}} & $\delta$=0.06 & Negligible & 0.42 \\
 &  & Integrated Gradients & \adjp{0.01}{*} & $\delta$=-0.10 & Negligible & 0.81 \\
 &  & GradientSHAP & \adjp{0.01}{*} & $\delta$=-0.09 & Negligible & 0.74 \\
 &  & Grad-CAM & \adjp{<0.01}{*} & $\delta$=-0.28 & Small & 1.00 \\
 &  & Feature Permutation & \adjp{<0.01}{*} & $\delta$=0.54 & Large & 1.00 \\

\bottomrule
\end{tabular}
\end{subtable}
\end{table}

\begin{table}[ht!]\ContinuedFloat
\centering
\small
\begin{subtable}[t]{\textwidth}
\centering
\caption{Oxford-IIIT Pet Dataset}
\label{tab:stats_all_b}

\begin{tabular}{lllcccc}
\toprule
Group & Metric & Method & $p$-value (BH-adj.) & Effect Size & Interp. & Power \\
\midrule
\multirow{10}{*}{ResNet Scale} &
\multirow{5}{*}{Relevance Rank Accuracy} &
Saliency & \adjp{0.62}{\phantom{*}} & $\eta^2$=0.00 & Negligible & 0.24 \\
& & Integrated Gradients & \adjp{0.86}{\phantom{*}} & $\eta^2$=0.00 & Negligible & 0.15 \\
& & GradientSHAP & \adjp{0.62}{\phantom{*}} & $\eta^2$=0.00 & Negligible & 0.21 \\
& & Grad-CAM & \adjp{<0.01}{*} & $\eta^2$=0.01 & Small & 0.93 \\
& & Feature Permutation & \adjp{0.44}{\phantom{*}} & $\eta^2$=0.00 & Negligible & 0.34 \\
\cmidrule(lr){2-7}
& \multirow{5}{*}{Dual Polarity Precision} &
Saliency & \adjp{0.47}{\phantom{*}} & $\eta^2$=0.00 & Negligible & 0.30 \\
& & Integrated Gradients & \adjp{0.32}{\phantom{*}} & $\eta^2$=0.00 & Negligible & 0.40 \\
& & GradientSHAP & \adjp{0.37}{\phantom{*}} & $\eta^2$=0.00 & Negligible & 0.35 \\
& & Grad-CAM & \adjp{0.01}{*} & $\eta^2$=0.01 & Small & 0.90 \\
& & Feature Permutation & \adjp{0.01}{*} & $\eta^2$=0.00 & Negligible & 0.88 \\
\midrule

\multirow{10}{*}{DenseNet Scale} &
\multirow{5}{*}{Relevance Rank Accuracy} &
Saliency & \adjp{0.91}{\phantom{*}} & $\eta^2$=0.00 & Negligible & 0.07 \\
& & Integrated Gradients & \adjp{0.97}{\phantom{*}} & $\eta^2$=0.00 & Negligible & 0.06 \\
& & GradientSHAP & \adjp{1.00}{\phantom{*}} & $\eta^2$=0.00 & Negligible & 0.05 \\
& & Grad-CAM & \adjp{0.09}{\phantom{*}} & $\eta^2$=0.00 & Negligible & 0.65 \\
& & Feature Permutation & \adjp{0.04}{*} & $\eta^2$=0.00 & Negligible & 0.72 \\
\cmidrule(lr){2-7}
& \multirow{5}{*}{Dual Polarity Precision} &
Saliency & \adjp{0.86}{\phantom{*}} & $\eta^2$=0.00 & Negligible & 0.11 \\
& & Integrated Gradients & \adjp{0.07}{\phantom{*}} & $\eta^2$=0.00 & Negligible & 0.67 \\
& & GradientSHAP & \adjp{0.10}{\phantom{*}} & $\eta^2$=0.00 & Negligible & 0.58 \\
& & Grad-CAM & \adjp{0.29}{\phantom{*}} & $\eta^2$=0.00 & Negligible & 0.35 \\
& & Feature Permutation & \adjp{<0.01}{*} & $\eta^2$=0.01 & Small & 1.00 \\
\midrule

\multirow{10}{*}{ViT (pretrained) Scale} &
\multirow{5}{*}{Relevance Rank Accuracy} &
Saliency & \adjp{0.04}{*} & $\delta$=0.08 & Negligible & 0.69 \\
& & Integrated Gradients & \adjp{0.01}{*} & $\delta$=0.10 & Negligible & 0.86 \\
& & GradientSHAP & \adjp{0.01}{*} & $\delta$=0.10 & Negligible & 0.87 \\
& & Grad-CAM & \adjp{0.87}{\phantom{*}} & $\delta$=-0.01 & Negligible & 0.05 \\
& & Feature Permutation & \adjp{0.26}{\phantom{*}} & $\delta$=0.05 & Negligible & 0.32 \\
\cmidrule(lr){2-7}
& \multirow{5}{*}{Dual Polarity Precision} &
Saliency & \adjp{0.10}{\phantom{*}} & $\delta$=0.07 & Negligible & 0.51 \\
& & Integrated Gradients & \adjp{0.32}{\phantom{*}} & $\delta$=0.04 & Negligible & 0.23 \\
& & GradientSHAP & \adjp{0.23}{\phantom{*}} & $\delta$=0.05 & Negligible & 0.32 \\
& & Grad-CAM & \adjp{0.66}{\phantom{*}} & $\delta$=-0.02 & Negligible & 0.09 \\
& & Feature Permutation & \adjp{0.31}{\phantom{*}} & $\delta$=0.04 & Negligible & 0.27 \\
\midrule

\multirow{10}{*}{ResNet Pretraining} &
\multirow{5}{*}{Relevance Rank Accuracy} &
Saliency & \adjp{<0.01}{*} & $\delta$=-0.13 & Negligible & 0.97 \\
& & Integrated Gradients & \adjp{0.09}{\phantom{*}} & $\delta$=-0.07 & Negligible & 0.53 \\
& & GradientSHAP & \adjp{0.09}{\phantom{*}} & $\delta$=-0.07 & Negligible & 0.54 \\
& & Grad-CAM & \adjp{<0.01}{*} & $\delta$=-0.20 & Small & 1.00 \\
& & Feature Permutation & \adjp{<0.01}{*} & $\delta$=-0.13 & Negligible & 0.97 \\
\cmidrule(lr){2-7}
& \multirow{5}{*}{Dual Polarity Precision} &
Saliency & \adjp{0.32}{\phantom{*}} & $\delta$=-0.04 & Negligible & 0.29 \\
& & Integrated Gradients & \adjp{0.82}{\phantom{*}} & $\delta$=-0.01 & Negligible & 0.11 \\
& & GradientSHAP & \adjp{0.88}{\phantom{*}} & $\delta$=-0.01 & Negligible & 0.08 \\
& & Grad-CAM & \adjp{<0.01}{*} & $\delta$=-0.21 & Small & 1.00 \\
& & Feature Permutation & \adjp{0.88}{\phantom{*}} & $\delta$=-0.01 & Negligible & 0.06 \\
\midrule

\multirow{10}{*}{DenseNet Pretraining\textsuperscript{$\dagger$}} &
\multirow{5}{*}{Relevance Rank Accuracy} &
Saliency & \adjp{<0.01}{*} & $\delta$=0.11 & Negligible & 0.92 \\
& & Integrated Gradients & \adjp{<0.01}{*} & $\delta$=0.11 & Negligible & 0.92 \\
& & GradientSHAP & \adjp{0.01}{*} & $\delta$=0.10 & Negligible & 0.87 \\
& & Grad-CAM & \adjp{0.15}{\phantom{*}} & $\delta$=0.06 & Negligible & 0.43 \\
& & Feature Permutation & \adjp{0.09}{\phantom{*}} & $\delta$=-0.07 & Negligible & 0.55 \\
\cmidrule(lr){2-7}
& \multirow{5}{*}{Dual Polarity Precision} &
Saliency & \adjp{0.25}{\phantom{*}} & $\delta$=0.05 & Negligible & 0.28 \\
& & Integrated Gradients & \adjp{<0.01}{*} & $\delta$=0.13 & Negligible & 0.98 \\
& & GradientSHAP & \adjp{<0.01}{*} & $\delta$=0.13 & Negligible & 0.97 \\
& & Grad-CAM & \adjp{0.15}{\phantom{*}} & $\delta$=0.06 & Negligible & 0.42 \\
& & Feature Permutation & \adjp{<0.01}{*} & $\delta$=-0.20 & Small & 1.00 \\

\bottomrule
\end{tabular}
\end{subtable}
\end{table}

\begin{table}[ht!]\ContinuedFloat
\centering
\small
\begin{subtable}[t]{\textwidth}
\centering
\caption{Chest X Pneumothorax Dataset}
\label{tab:stats_all_c}

\begin{tabular}{lllcccc}
\toprule
Group & Metric & Method & $p$-value (BH-adj.) & Effect Size & Interp. & Power \\
\midrule
\multirow{10}{*}{ResNet Scale} &
\multirow{5}{*}{Relevance Rank Accuracy} &
Saliency & \adjp{<0.01}{*} & $\eta^2$=0.01 & Small & 0.99 \\
& & Integrated Gradients & \adjp{<0.01}{*} & $\eta^2$=0.02 & Small & 1.00 \\
& & GradientSHAP & \adjp{<0.01}{*} & $\eta^2$=0.01 & Small & 1.00 \\
& & Grad-CAM & \adjp{0.14}{\phantom{*}} & $\eta^2$=0.00 & Negligible & 0.57 \\
& & Feature Permutation & \adjp{0.01}{*} & $\eta^2$=0.00 & Negligible & 0.88 \\
\cmidrule(lr){2-7}
& \multirow{5}{*}{Dual Polarity Precision} &
Saliency & \adjp{<0.01}{*} & $\eta^2$=0.00 & Negligible & 0.99 \\
& & Integrated Gradients & \adjp{<0.01}{*} & $\eta^2$=0.02 & Small & 1.00 \\
& & GradientSHAP & \adjp{<0.01}{*} & $\eta^2$=0.02 & Small & 1.00 \\
& & Grad-CAM & \adjp{0.05}{\phantom{*}} & $\eta^2$=0.00 & Negligible & 0.72 \\
& & Feature Permutation & \adjp{<0.01}{*} & $\eta^2$=0.02 & Small & 1.00 \\
\midrule

\multirow{10}{*}{DenseNet Scale} &
\multirow{5}{*}{Relevance Rank Accuracy} &
Saliency & \adjp{0.31}{\phantom{*}} & $\eta^2$=0.00 & Negligible & 0.34 \\
& & Integrated Gradients & \adjp{0.90}{\phantom{*}} & $\eta^2$=0.00 & Negligible & 0.06 \\
& & GradientSHAP & \adjp{0.73}{\phantom{*}} & $\eta^2$=0.00 & Negligible & 0.11 \\
& & Grad-CAM & \adjp{0.07}{\phantom{*}} & $\eta^2$=0.00 & Negligible & 0.66 \\
& & Feature Permutation & \adjp{0.08}{\phantom{*}} & $\eta^2$=0.00 & Negligible & 0.60 \\
\cmidrule(lr){2-7}
& \multirow{5}{*}{Dual Polarity Precision} &
Saliency & \adjp{0.73}{\phantom{*}} & $\eta^2$=0.00 & Negligible & 0.13 \\
& & Integrated Gradients & \adjp{0.04}{*} & $\eta^2$=0.00 & Negligible & 0.73 \\
& & GradientSHAP & \adjp{0.05}{\phantom{*}} & $\eta^2$=0.00 & Negligible & 0.71 \\
& & Grad-CAM & \adjp{0.36}{\phantom{*}} & $\eta^2$=0.00 & Negligible & 0.28 \\
& & Feature Permutation & \adjp{<0.01}{*} & $\eta^2$=0.02 & Small & 1.00 \\
\midrule

\multirow{10}{*}{ViT (pretrained) Scale} &
\multirow{5}{*}{Relevance Rank Accuracy} &
Saliency & \adjp{0.02}{*} & $\delta$=0.09 & Negligible & 0.77 \\
& & Integrated Gradients & \adjp{0.73}{\phantom{*}} & $\delta$=0.01 & Negligible & 0.07 \\
& & GradientSHAP & \adjp{0.96}{\phantom{*}} & $\delta$=0.00 & Negligible & 0.04 \\
& & Grad-CAM & \adjp{<0.01}{*} & $\delta$=-0.17 & Small & 1.00 \\
& & Feature Permutation & \adjp{0.24}{\phantom{*}} & $\delta$=-0.04 & Negligible & 0.30 \\
\cmidrule(lr){2-7}
& \multirow{5}{*}{Dual Polarity Precision} &
Saliency & \adjp{0.34}{\phantom{*}} & $\delta$=-0.04 & Negligible & 0.17 \\
& & Integrated Gradients & \adjp{0.01}{*} & $\delta$=0.10 & Negligible & 0.83 \\
& & GradientSHAP & \adjp{0.16}{\phantom{*}} & $\delta$=0.06 & Negligible & 0.37 \\
& & Grad-CAM & \adjp{0.19}{\phantom{*}} & $\delta$=-0.05 & Negligible & 0.37 \\
& & Feature Permutation & \adjp{0.08}{\phantom{*}} & $\delta$=-0.07 & Negligible & 0.55 \\
\midrule

\multirow{10}{*}{ResNet Pretraining} &
\multirow{5}{*}{Relevance Rank Accuracy} &
Saliency & \adjp{0.07}{\phantom{*}} & $\delta$=-0.07 & Negligible & 0.57 \\
& & Integrated Gradients & \adjp{0.01}{*} & $\delta$=-0.09 & Negligible & 0.84 \\
& & GradientSHAP & \adjp{0.08}{\phantom{*}} & $\delta$=-0.06 & Negligible & 0.56 \\
& & Grad-CAM & \adjp{0.77}{\phantom{*}} & $\delta$=-0.01 & Negligible & 0.07 \\
& & Feature Permutation & \adjp{0.60}{\phantom{*}} & $\delta$=-0.02 & Negligible & 0.14 \\
\cmidrule(lr){2-7}
& \multirow{5}{*}{Dual Polarity Precision} &
Saliency & \adjp{0.01}{*} & $\delta$=-0.10 & Negligible & 0.81 \\
& & Integrated Gradients & \adjp{0.88}{\phantom{*}} & $\delta$=0.01 & Negligible & 0.06 \\
& & GradientSHAP & \adjp{0.15}{\phantom{*}} & $\delta$=0.06 & Negligible & 0.40 \\
& & Grad-CAM & \adjp{0.80}{\phantom{*}} & $\delta$=0.01 & Negligible & 0.05 \\
& & Feature Permutation & \adjp{0.60}{\phantom{*}} & $\delta$=0.02 & Negligible & 0.12 \\
\midrule

\multirow{10}{*}{DenseNet Pretraining} &
\multirow{5}{*}{Relevance Rank Accuracy} &
Saliency & \adjp{<0.01}{*} & $\delta$=0.31 & Small & 1.00 \\
& & Integrated Gradients & \adjp{<0.01}{*} & $\delta$=0.15 & Small & 0.99 \\
& & GradientSHAP & \adjp{<0.01}{*} & $\delta$=0.16 & Small & 1.00 \\
& & Grad-CAM & \adjp{0.31}{\phantom{*}} & $\delta$=0.03 & Negligible & 0.21 \\
& & Feature Permutation & \adjp{0.08}{\phantom{*}} & $\delta$=-0.05 & Negligible & 0.57 \\
\cmidrule(lr){2-7}
& \multirow{5}{*}{Dual Polarity Precision} &
Saliency & \adjp{0.01}{*} & $\delta$=0.11 & Negligible & 0.88 \\
& & Integrated Gradients & \adjp{0.60}{\phantom{*}} & $\delta$=0.02 & Negligible & 0.10 \\
& & GradientSHAP & \adjp{0.60}{\phantom{*}} & $\delta$=0.02 & Negligible & 0.10 \\
& & Grad-CAM & \adjp{0.24}{\phantom{*}} & $\delta$=0.05 & Negligible & 0.28 \\
& & Feature Permutation & \adjp{0.25}{\phantom{*}} & $\delta$=-0.05 & Negligible & 0.29 \\

\bottomrule
\end{tabular}

\end{subtable}
\end{table}

\clearpage
\section{Sanity Check Examples}
\label{app:examples_sanity_checks}
Additional examples of conducted sanity checks (layer-wise randomization and input perturbation).

\begin{figure*}[h!]
\centering
\includegraphics[width=0.89\textwidth]{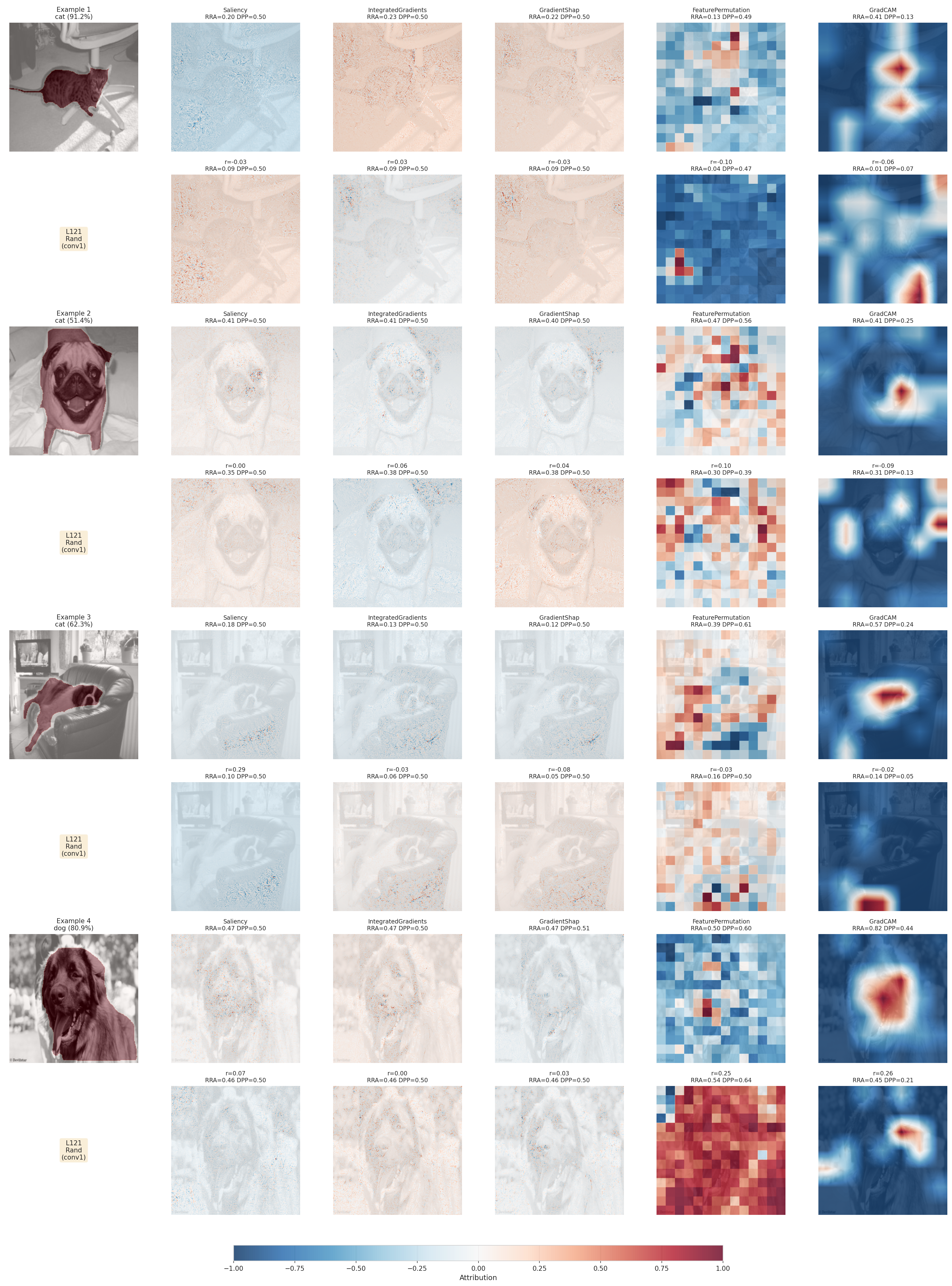}
\caption{Layer-wise randomisation of ImageNet-pretrained ResNet-50.}
\label{layer_wise_random}
\end{figure*}

\begin{figure*}[h!]
\centering
\includegraphics[width=0.89\textwidth]{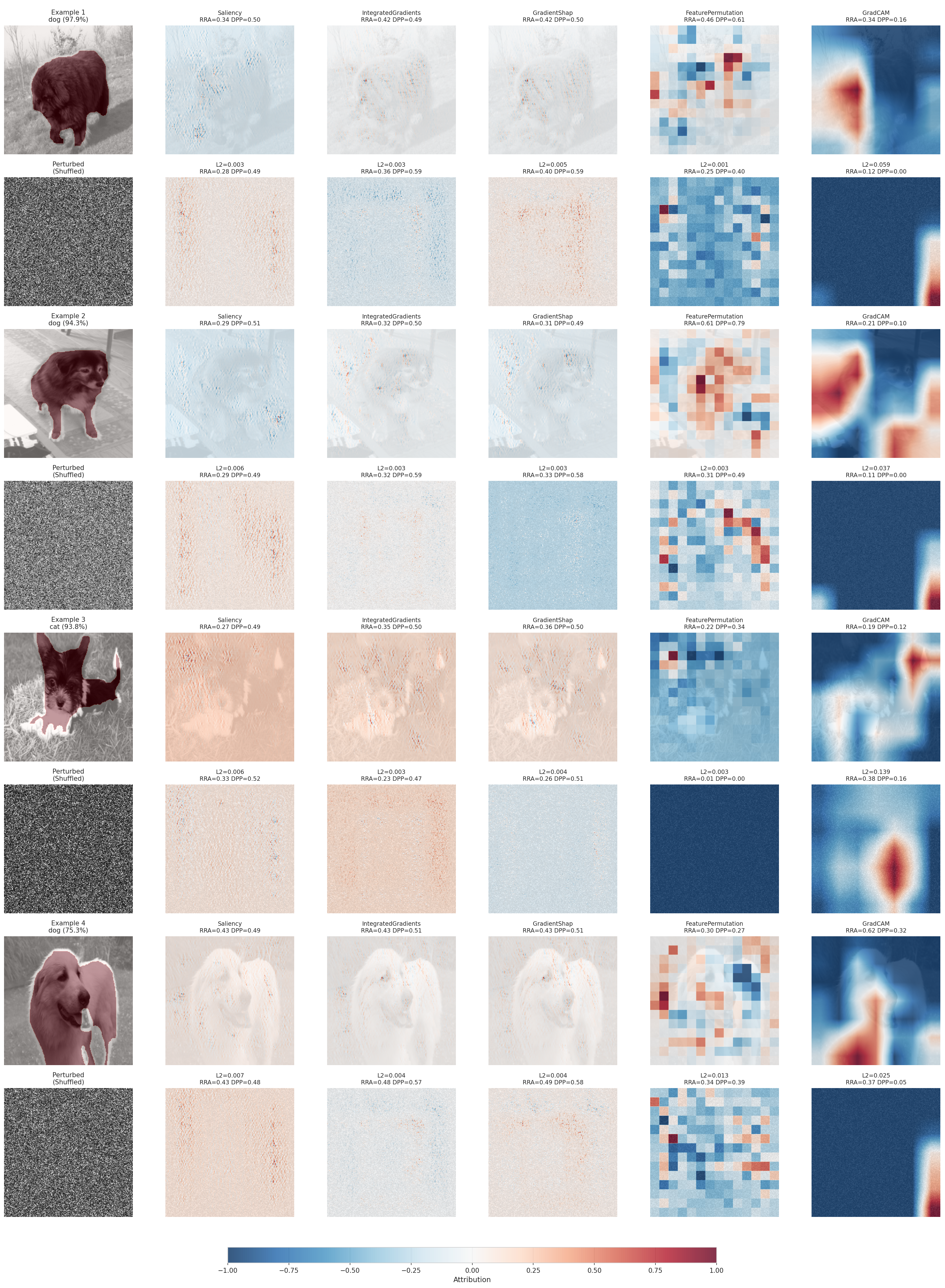}
\caption{Input-wise randomisation of ResNet-18.}
\label{input_random}
\end{figure*}

\section{Examples of Explanations on Chest X-ray Pneumothorax dataset}
\label{app:pneumo_images_appendix}

Below, we present additional qualitative examples of XAI explanations generated on the Chest X-ray Pneumothorax dataset. Results are shown for two model variants: a ResNet-50 pretrained on ImageNet and a DenseNet-121 pretrained on CheXpert.

\begin{figure*}[h!]
\centering
\begin{subfigure}{\textwidth}
    \centering
    \caption{ResNet-50 (IN-1k pretrained), examples 1--4.}
    \includegraphics[width=0.875\textwidth]{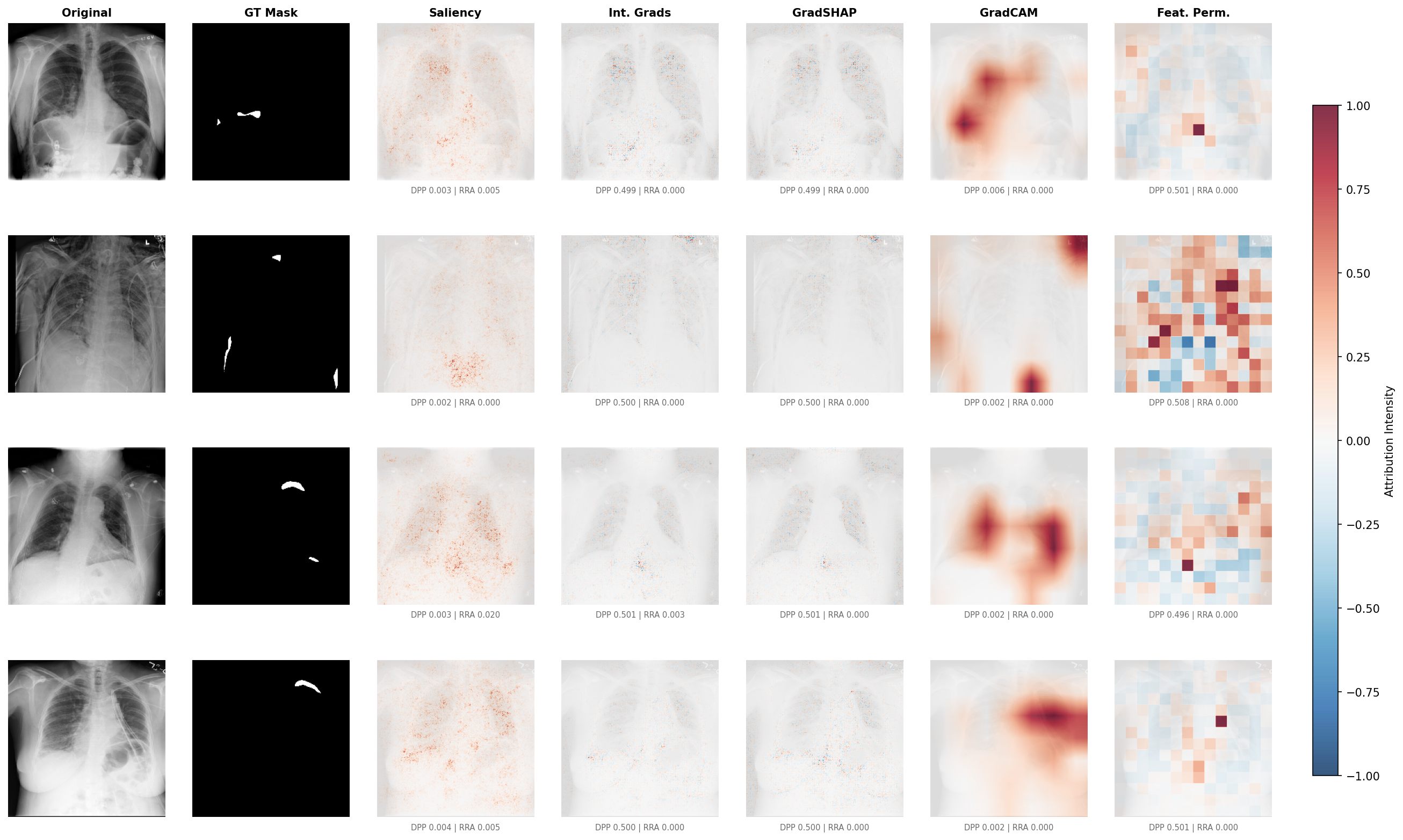}
    \label{fig:pneumo_resnet50_1}
\end{subfigure}

\begin{subfigure}{\textwidth}
    \centering
    \caption{ResNet-50 (IN-1k pretrained), examples 5--8.}
    \includegraphics[width=0.875\textwidth]{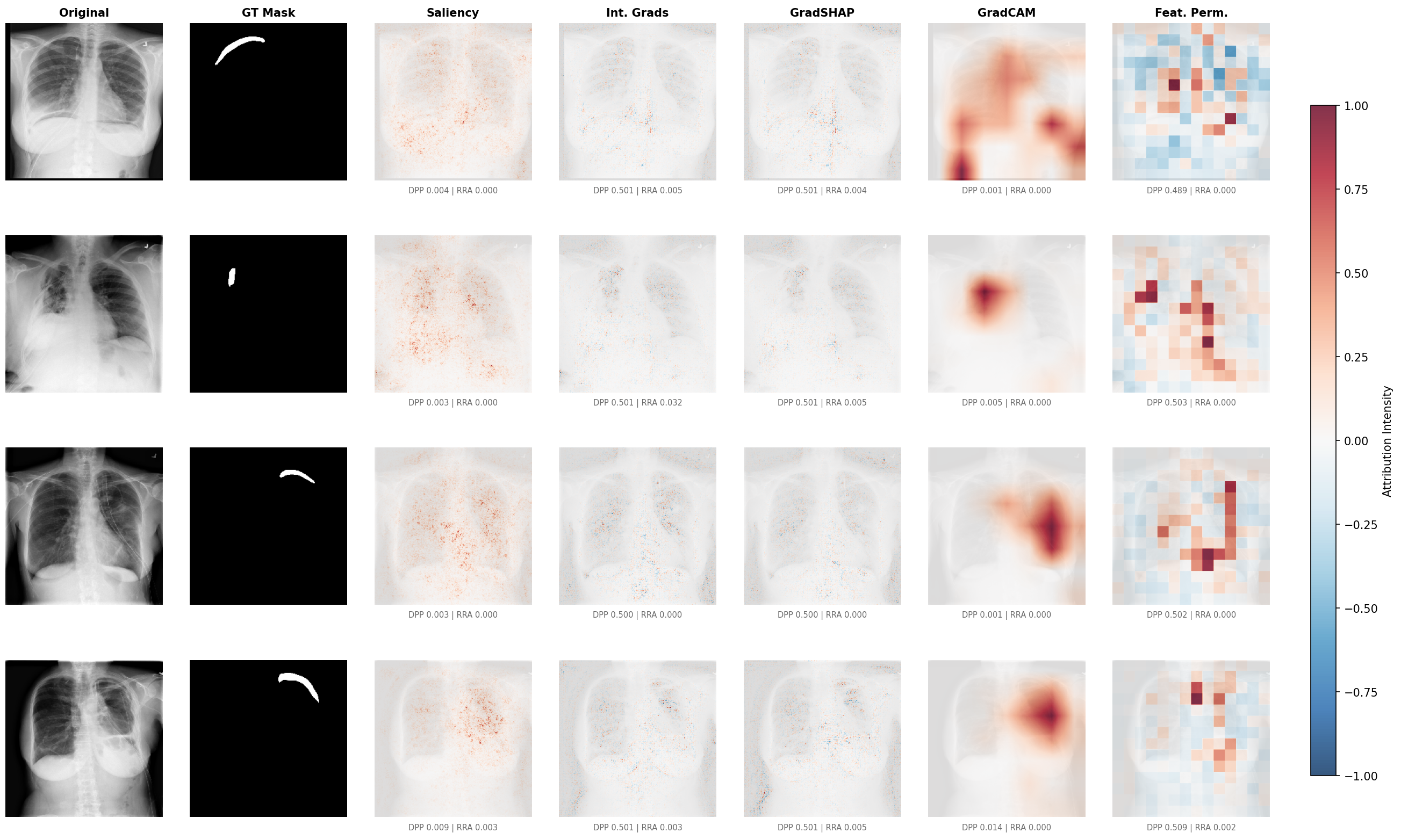}
    \label{fig:pneumo_resnet50_2}
\end{subfigure}
\caption{Saliency-based explanations on the Chest X-ray Pneumothorax dataset using a ResNet-50 pretrained on ImageNet. Each row shows the original X-ray, the ground truth mask, and attributions from five methods: Saliency, Integrated Gradients, GradientSHAP, Grad-CAM, and Feature Permutation. DPP and RRA scores are reported below each attribution map.}
\label{fig:pneumo_resnet50}
\end{figure*}

\begin{figure*}[h!]
\centering
\begin{subfigure}{\textwidth}
    \centering
    \caption{DenseNet-121 (CheXpert pretrained), examples 9--12.}
    \includegraphics[width=0.875\textwidth]{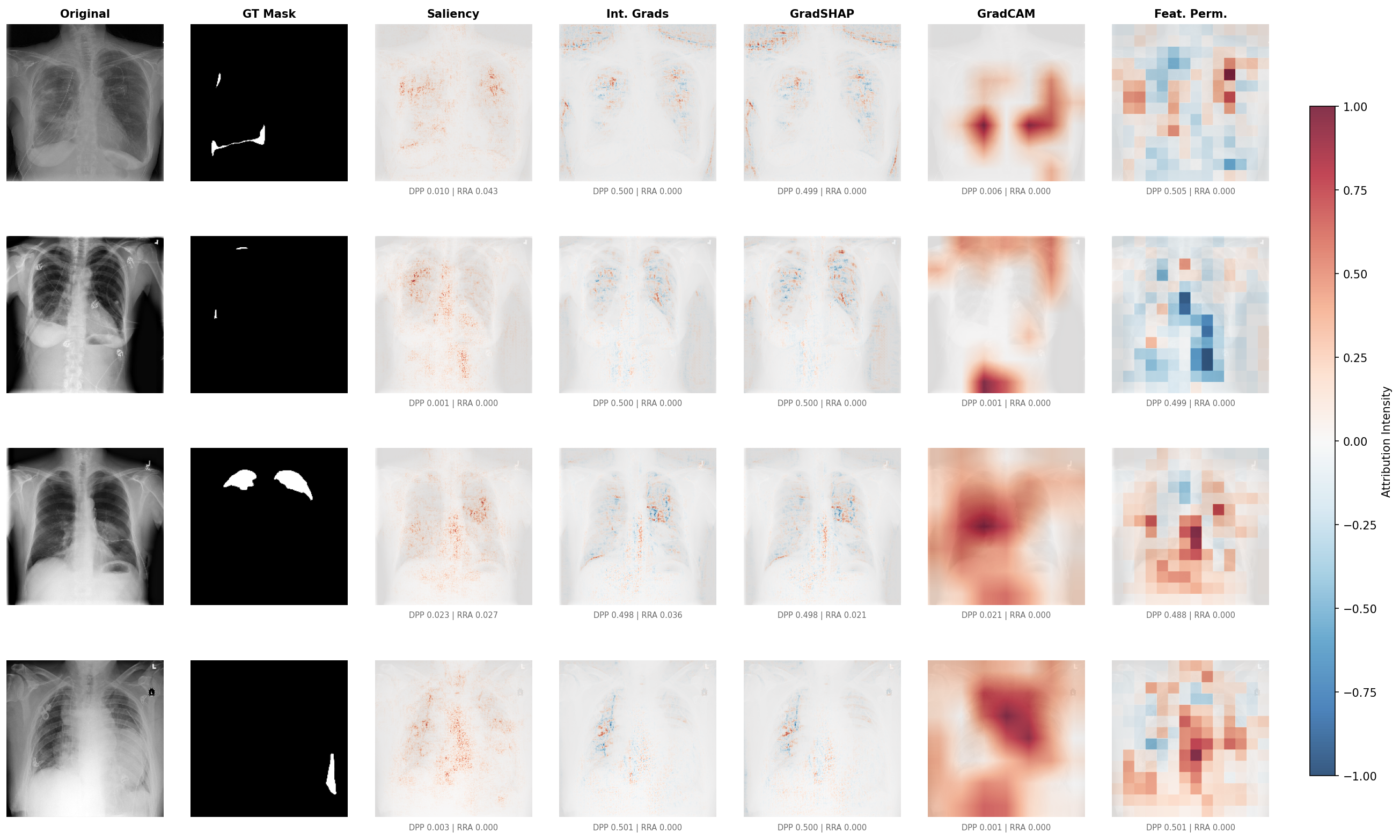}
    \label{fig:pneumo_densenet121_1}
\end{subfigure}

\begin{subfigure}{\textwidth}
    \centering
    \caption{DenseNet-121 (CheXpert pretrained), examples 13--16.}
    \includegraphics[width=0.875\textwidth]{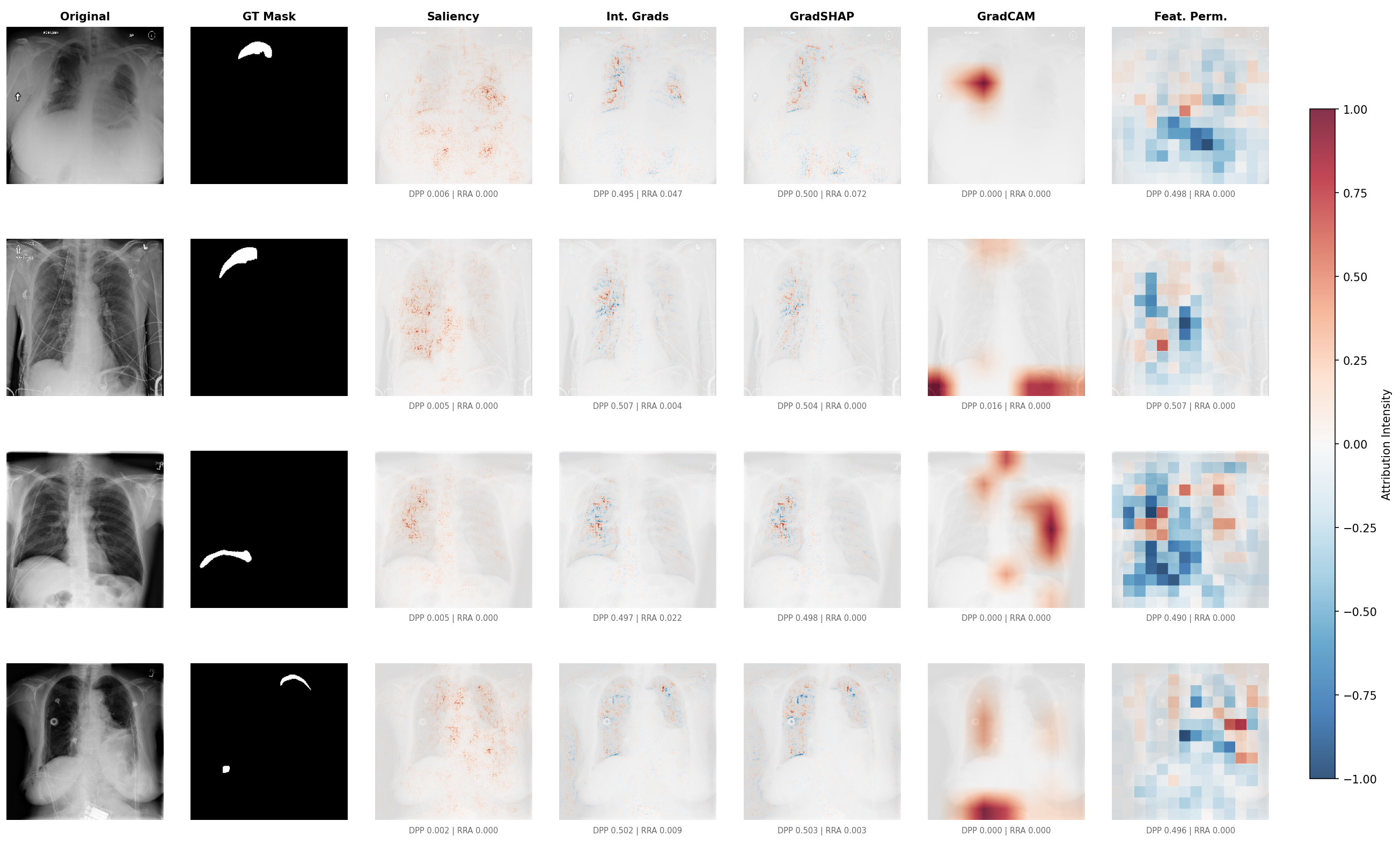}
    \label{fig:pneumo_densenet121_2}
\end{subfigure}
\caption{Saliency-based explanations on the Chest X-ray Pneumothorax dataset using a DenseNet-121 pretrained on CheXpert. Each row shows the original X-ray, the ground truth mask, and attributions from five methods: Saliency, Integrated Gradients, GradientSHAP, Grad-CAM, and Feature Permutation. DPP and RRA scores are reported below each attribution map.}
\label{fig:pneumo_densenet121}
\end{figure*}

\end{document}